\documentclass[twoside,11pt]{article}

\usepackage{jmlr2e}
\usepackage{graphicx}
\usepackage{booktabs}
\usepackage{subfigure}
\usepackage{amsfonts}
\usepackage{amsmath}
\usepackage{algorithm}
\usepackage{algorithmic}
\usepackage{caption}
\usepackage{tabularx}
\usepackage{amsmath}
\usepackage{epstopdf}
\usepackage{verbatim}
\usepackage{color}
\usepackage{multirow}
\usepackage{url}

\newcommand{\calM}{{\cal M}}
\newcommand{\calJ}{{\cal J}}
\newcommand{\calL}{{\cal L}}
\newcommand{\calP}{q} 

\newcommand{\s}[1]{^{(#1)}}

\newcommand{\is}[4]{\pi^{(#1),#2}_{#3_#4}}
\newcommand{\st}[5]{a^{(#1),#2}_{#3_{#4},#3_{#5}}}

\newcommand{\Normal}[2]{{\cal N}(#1,#2)}
\newcommand{\argmax}{ \mathop{\mathrm{argmax}} }
\newcommand{\EV}{\mathrm{E}}
\newcommand{\Id}{\mathrm{I}}

\newcommand{\refeqn}[1]{(\ref{#1})}

\newcommand{\eg}{e.g., }

\newcommand{\beqa}{\begin{eqnarray}}
\newcommand{\eeqa}{\end{eqnarray}}
\newcommand{\beq}{\begin{equation}}
\newcommand{\eeq}{\end{equation}}

\newcommand{\subgm}{_{\text{\tiny GMM}}}

\newcommand{\subhmm}{_{\text{\tiny HMM}}}

\newcommand{\gnd}[1]{}{}  

\definecolor{mygreen}{rgb}{0,0.7,0}
\newcommand{\ted}[1]{#1}  
\newcommand{\tn}[1]{\textcolor{mygreen}{[Ton: #1]}}  

\newcommand{\comments}[1]{}

\jmlrheading{1}{2012}{1-44}{10/12}{??/??}{Emanuele Coviello}

\ShortHeadings{Clustering HMMs with variational HEM}{Coviello, Chan, Lanckriet}
\firstpageno{1}

\begin{document}

\title{Clustering hidden Markov models with variational HEM
}

\author{\name Emanuele Coviello \email ecoviell@ucsd.edu \\
       \addr Department of Electrical and Computer Engineering\\
       University of California, San Diego\\
       La Jolla, CA 92093, USA
       \AND
\name Antoni B. Chan \email abchan@cityu.edu.hk \\
       \addr Department of Computer Science\\
       City University of Hong Kong\\
      Kowloon Tong, Hong Kong
       \AND
\name Gert R.G. Lanckriet \email gert@ece.ucsd.edu \\
       \addr Department of Electrical and Computer Engineering\\
       University of California, San Diego\\
       La Jolla, CA 92093, USA
       }

\editor{Editor Name}

\maketitle

\begin{abstract}
The hidden Markov model (HMM) is a widely-used generative model that copes with sequential data, assuming that each observation is conditioned on the state of a hidden Markov chain.
In this paper, we derive a novel algorithm to cluster HMMs based on the hierarchical EM (HEM) algorithm.
The proposed algorithm i) clusters a given collection of HMMs into groups of HMMs that are similar, in terms of the distributions they represent, and ii) characterizes each group by a ``cluster center'', i.e., a novel HMM that is representative for the group, in a manner that is consistent with the underlying generative model of the HMM.
To cope with intractable inference in the E-step, the HEM algorithm is formulated as a variational optimization problem, and efficiently solved for the HMM case by leveraging an appropriate variational approximation.
The benefits of the proposed algorithm, which we call variational HEM (VHEM), are demonstrated on several tasks involving time-series data, such as hierarchical clustering of motion capture sequences, and automatic annotation and retrieval of music and of online hand-writing data, showing improvements over current methods.
In particular, our variational HEM algorithm 
effectively leverages large amounts of data when learning annotation models by using an efficient hierarchical estimation procedure, which reduces learning times and memory requirements, while improving model robustness through better regularization.
\end{abstract}


\section{Introduction}

The hidden Markov model (HMM) \citep{Rabiner93} is a probabilistic model that assumes a signal is generated by a double embedded stochastic process. A discrete-time hidden state process, which evolves as a Markov chain, encodes the dynamics of the signal, while an observation process encodes the appearance of the signal at each time, conditioned on the current state.
HMMs have been successfully applied to a variety of fields, including speech recognition~\citep{Rabiner93}, music analysis~\citep{qi2007music} and identification~\citep{batlle2002automatic}, online hand-writing recognition~\citep{nag1986script}, analysis of biological sequences~\citep{krogh1994hidden}, 
and clustering of time series data~\citep{jebara2007, smyth1997}.

This paper is about clustering HMMs. More precisely, we are interested in an algorithm that, given a collection of HMMs, partitions them into $K$ clusters of ``similar'' HMMs, while also learning a representative HMM ``cluster center'' that concisely and appropriately represents each cluster. This is similar to standard k-means clustering, except that the data points are HMMs now instead of vectors in $\mathbb{R}^d$.

Various applications motivate the design of HMM clustering algorithms, ranging from
hierarchical clustering of sequential data (e.g., speech 
or motion sequences modeled by HMMs ~\citep{jebara2007}),
to hierarchical indexing for fast retrieval, to reducing the computational complexity of estimating mixtures of HMMs from large datasets (e.g., semantic annotation models for music and video) 
~---~by clustering HMMs, efficiently estimated from many small subsets of the data, into a more compact mixture model of all data.
However, there has been little work on HMM clustering and, therefore, its applications.

Existing approaches to clustering HMMs operate directly on the HMM \emph{parameter} space, by grouping HMMs according to a suitable pairwise distance defined in terms of the HMM parameters.
However, as HMM parameters lie on a non-linear manifold, 
a simple application of the k-means algorithm will not succeed in the task, since it assumes real vectors in a Euclidean space. 
In addition, such an approach would have the additional complication that HMM parameters for a particular generative model are not unique, i.e., a permutation of the states leads to the same generative model.
One solution, proposed by \cite{jebara2007}, first constructs an appropriate similarity matrix between all HMMs that are to be clustered (e.g., based on the Bhattacharya affinity, which depends non-linearly on the HMM parameters \citep{jebara2004probability}),
and then applies spectral clustering. While this approach has proven successful to group HMMs into similar clusters~\citep{jebara2007}, it does 
\ted{not directly address the issue of generating HMM cluster centers.}
Each cluster can still be represented by choosing one of the given HMMs, e.g., the HMM which the spectral clustering procedure maps the closest to each spectral clustering center.
However, this may be suboptimal for some applications of HMM clustering, for example in hierarchical estimation of annotation models. 
Another distance between HMM distributions suitable for spectral clustering is the KL divergence, which in practice has been approximated by sampling sequences from one model and computing their log-likelihood under the other \citep{juang1985,zhong2003unified,yin2005integrating}.

Instead, in this paper we propose to cluster HMMs \emph{directly} with respect to the \emph{probability distributions} they represent. The probability distributions of the HMMs are used throughout the whole clustering algorithm, and not only to construct an initial embedding as in \citep{jebara2007}.
By clustering the output distributions of the HMMs, marginalized over the hidden-state distributions, we avoid the issue of multiple equivalent parameterizations of the hidden states.
We derive a hierarchical expectation maximization (HEM) algorithm that, starting from a collection of input HMMs, estimates a smaller mixture model of HMMs that concisely represents and clusters the input HMMs (i.e., the input HMM distributions guide the estimation of the output mixture distribution).

The HEM algorithm is a \emph{generalization} of the EM algorithm~---~the EM algorithm can be considered as a special case of HEM for a mixture of delta functions as input. The main difference between HEM and EM is in the E-step. While the EM algorithm computes the sufficient statistics given the observed data, the HEM algorithm calculates the expected sufficient statistics averaged over all possible observations generated by the input probability models.
Historically, the first HEM algorithm was designed to cluster \emph{Gaussian} probability distributions \citep{Vasc1998}.
This algorithm starts from a Gaussian mixture model (GMM) with $K\s{b}$ components and reduces it to another GMM with fewer components, where each of the mixture components of the reduced GMM represents, i.e., \emph{clusters}, a group of the original Gaussian mixture components. More recently, \cite{Chan2010tr} derived an HEM algorithm to cluster  \emph{dynamic texture} (DT) models (i.e., linear dynamical systems, LDSs) through their probability distributions.
HEM has been applied successfully to construct GMM hierarchies for efficient image indexing~\citep{Vasc2001b},
to cluster video represented by DTs~\citep{Chan2010cvpr}, and to estimate GMMs or DT mixtures (DTMs, i.e., LDS mixtures) from large datasets for semantic annotation of images~\citep{carneiro07}, video~\citep{Chan2010cvpr} and music \citep{music:turnbull08,coviello2011}.

To extend the HEM framework for GMMs to mixtures of HMMs (or hidden Markov mixture models, H3Ms), additional marginalization over the hidden-state processes is required, as with DTMs. However, while Gaussians and DTs allow tractable inference in the E-step of HEM, this is no longer the case for HMMs. Therefore, in this work, we derive a variational formulation of the HEM algorithm (VHEM), and then leverage a variational \emph{approximation} derived by \cite{hershey2008variational} (which has not been used in a learning context so far) to make the inference in the E-step tractable. 
The resulting algorithm not only 
clusters HMMs, but also learns novel HMMs that are representative centers of each cluster. 
The resulting VHEM algorithm can be generalized to handle other classes of graphical models, for which exact computation of the E-step in the standard HEM would be intractable, by leveraging similar variational approximations~---~e.g., the more general case of HMMs with emission probabilities that are (mixtures of) continuous exponential family distributions.

Compared to the spectral clustering algorithm of \citep{jebara2007}, 
the VHEM algorithm has several advantages that make it suitable for a variety of applications. 
 First, the VHEM algorithm is capable of both clustering, as well as learning \emph{novel} cluster centers, in a manner that is consistent with the underlying generative probabilistic framework.
In addition, since it does not require sampling steps, it is 
also scalable with low memory requirements.
As a consequence, VHEM for HMMs allows for efficient estimation of HMM mixtures from large datasets using a hierarchical estimation procedure.
In particular, intermediate HMM mixtures are first estimated in \emph{parallel} by running the EM algorithm on small independent portions of the dataset. The final model is then estimated from the intermediate models using the VHEM algorithm.
Because VHEM is based on maximum-likelihood principles, it drives model estimation towards similar optimal parameter values as performing maximum-likelihood estimation on the full dataset.
In addition, by averaging over all possible observations compatible with the input models in the E-step, VHEM provides an implicit form of regularization that prevents over-fitting and improves robustness of the learned models, compared to a direct application of the EM algorithm on the full dataset.
Note that, in contrast to \citep{jebara2007}, VHEM does not construct a kernel embedding, which is a costly operation 
that 
requires calculating of $O((K\s{b})^2)$ pairwise similarity scores between all input HMMs to form the similarity matrix, as well as the inversion of this matrix.

In summary, the contributions of this paper are three-fold: 
i) we derive a variational formulation of the HEM algorithm for clustering HMMs, which generates novel HMM centers representative of each cluster;
ii) we evaluate VHEM on a variety of clustering, annotation, and retrieval problems involving time-series data, showing improvement over current clustering methods;
iii) we demonstrate in experiments that VHEM can effectively learn HMMs from large sets of data, more efficiently than standard EM, while improving model robustness through better regularization.
With respect to our previous work, the VHEM algorithm for H3M was originally proposed in \citep{coviello2012variational}

The remainder of the paper is organized as follows. We review the hidden Markov model (HMM) and the hidden Markov mixture model (H3M) in Section \ref{sec:hmm}. We present the derivation of the VHEM-H3M algorithm in Section \ref{sec:clustering}, followed by a discussion in Section \ref{sec:dis}.
Finally, we present experimental results in Sections \ref{sec:exp:cluster} and \ref{sec:exP:ann}.

\section{The hidden Markov (mixture) model}\label{sec:hmm}

A hidden Markov model (HMM) $\calM$ assumes a sequence of $\tau$ observations $y_{1:\tau}=\{y_1,\ldots,y_{\tau}\}$ is generated by a double embedded stochastic process, where each observation (or emission) $y_t$ at time $t$ depends on the state of a discrete hidden variable $x_t$, and the sequence of hidden states $x_{1:\tau}=\{x_1,\ldots,x_{\tau}\}$ evolves as a first-order Markov chain.
The hidden variables can take one of $S$ values, $\{1,\ldots,S\}$, and the evolution of the hidden process is
encoded in a state transition matrix $A = [a_{\beta,\beta'}]_{\beta,\beta' = 1,\dots,S}$, where each entry,
$a_{\beta,\beta'} = p(x_{t+1} = \beta' | x_t = \beta, \calM)$, is the probability of transitioning from state $\beta$ to state $\beta'$, and an initial state distribution $\pi = [\pi_1,\dots,\pi_S]$, where $\pi_\beta = p(x_1 = \beta| \calM)$.

Each state $\beta$ generates observations according to an emission probability density function, $p(y_t|x_t = \beta, \calM )$. Here, we assume the emission density is {\em time-invariant}, and modeled as a Gaussian mixture model (GMM) with $M$ components:
\begin{align}
p(y| x =\beta,\calM) &= \sum_{m = 1}^{M}  c_{\beta,m} p(y|\zeta = m,x=\beta,\calM),  
\label{eqn:GMM}
\end{align}
where 
$\zeta \sim \mathrm{multinomial} (c_{\beta,1},\dots,c_{\beta,M})$ is the hidden assignment variable that selects the mixture component,
 with $c_{\beta,m}$ as the mixture weight of the $m$th component, and 
 each component is a multivariate Gaussian distribution,
\begin{align}  
  p(y|\zeta = m,x=\beta,\calM) &= \Normal{y;\mu_{\beta,m}}{\Sigma_{\beta,m}} ,
\end{align}
with mean $\mu_{\beta,m}$ and covariance matrix $\Sigma_{\beta,m}$.
The HMM is specified by the parameters 
\begin{equation}
\calM = \{\pi, A, \{\{c_{\beta,m}, \mu_{\beta,m}, \Sigma_{\beta,m}\}_{m=1}^{M}\}_{\beta=1}^{S} \},
\end{equation}
 which can be efficiently learned from an observation sequence $y_{1:\tau}$ with the Baum-Welch algorithm~\citep{Rabiner93}, which is based on maximum likelihood estimation.

\ted{The probability distribution of a state sequence $x_{1:\tau}$ generated by an HMM $\calM$ is }
	\begin{align}
	p(x_{1:\tau}|\calM)  = p(x_1|\calM) \prod_{t=2}^{\tau} p(x_t|x_{t-1},\calM) = 
	\pi_{x_{1}} \prod_{t=2}^{\tau} a_{x_{t-1},x_{t}},
	\label{eqn:HMM-state}
	\end{align}
\ted{while the joint likelihood of an observation sequence $y_{1:\tau}$ and a state sequence $x_{1:\tau}$ is}
	\begin{align}
	p(y_{1:\tau},x_{1:\tau}|\calM) = p(y_{1:\tau}|x_{1:\tau},\calM) p(x_{1:\tau}|\calM) = 
	p(x_1|\calM) \prod_{t=2}^{\tau} p(x_t|x_{t-1},\calM) \prod_{t=1}^{\tau} p(y_t|x_t,\calM).
	\label{eqn:HMM-joint}
	\end{align}
\ted{Finally, the observation likelihood of $y_{1:\tau}$ is obtained by marginalizing out the state sequence from the joint likelihood,}
	\begin{align}
	p(y_{1:\tau}|\calM) = \sum_{x_{1:\tau}} p(y_{1:\tau},x_{1:\tau}|\calM) =  \sum_{x_{1:\tau}} p(y_{1:\tau}|x_{1:\tau},\calM) p(x_{1:\tau}|\calM),
	\label{eqn:HMM-obs}
	\end{align}
\ted{where the summation is over all state sequences of length $\tau$, and can be performed efficiently using the {\em forward algorithm} \citep{Rabiner93}.}



A hidden Markov mixture model (H3M) \citep{smyth1997} models a set of observation sequences as samples from a group of $K$ hidden Markov models, each associated to a specific sub-behavior.
For a given sequence, an assignment variable $z \sim \mathrm{multinomial}(\omega_1, \cdots, \omega_{K})$ selects the parameters of one of the $K$ HMMs, where the $k$th HMM is selected with probability $\omega_k$.
Each mixture component is parametrized by 
\begin{equation}
\calM_z = \{\pi^z, A^z, \{\{c_{\beta,m}^z, \mu_{\beta,m}^z, \Sigma_{\beta,m}^z\}_{m=1}^{M}\}_{\beta=1}^{S} \},
\end{equation} 
and the H3M is parametrized by $\calM = \{ \omega_z, \calM_z\}_{z=1}^{K}$, which can be estimated from a collection of observation sequences using the EM algorithm \citep{smyth1997}.


To reduce clutter, here we assume that all the HMMs have the same number $S$ of hidden states and that all emission probabilities have $M$ mixture components. Our derivation could be easily extended to the more general case though.

\section{Clustering hidden Markov models}\label{sec:clustering}

Algorithms for clustering HMMs can serve a wide range of applications, from 
hierarchical clustering of sequential data (e.g., speech or motion sequences modeled by HMMs ~\citep{jebara2007}),
to hierarchical indexing for fast retrieval, to reducing the computational complexity of estimating mixtures of HMMs from large weakly-annotated datasets
~---~by clustering HMMs, efficiently estimated from many small subsets of the data, into a more compact mixture model of all data.

In this work we derive a hierarchical EM algorithm for clustering HMMs (HEM-H3M)  with respect to their probability distributions.
We approach the problem of clustering HMMs 
as reducing an input HMM mixture with a large number of components to a new mixture with fewer components. Note that different HMMs in the input mixture are allowed to have different weights (i.e.,  the mixture weights $\{\omega_z\}_{z=1}^{K}$ are not necessarily all equal). 
%
%

\ted{
One method for estimating the reduced mixture model is to
generate samples from the input mixture, and then perform maximum likelihood estimation, i.e., maximize the log-likelihood of these samples.
However, to avoid explicitly generating these samples, we instead maximize the {\em expectation} of the log-likelihood with respect to the input mixture model, thus averaging over all possible samples from the input mixture model.
In this way, the dependency on the samples is replaced by a marginalization with respect to the input mixture model.
}
While such marginalization
is tractable for Gaussians and DTs, this is no longer the case for HMMs. Therefore, in this work, we i) derive a variational formulation of the HEM algorithm (VHEM), and ii) specialize it to the HMM case by leveraging a variational approximation proposed by \cite{hershey2008variational}.
Note that~\citep{hershey2008variational} was proposed as an alternative to MCMC sampling for the computation of the KL divergence between two HMMs, and has not been used in a learning context so far.

We present the problem formulation in Section \ref{subsec:form}, and derive the algorithm in Sections \ref{subsev:var_lb}, \ref{subsev:e:new} and \ref{subsev:m}.

\subsection{Formulation}\label{subsec:form}

Let $\calM\s{b}$ be a base hidden Markov mixture model with $K\s{b}$ components.
The goal of the VHEM algorithm is to find a reduced hidden Markov mixture model $\calM\s{r}$ with $K\s{r}<K\s{b}$ (i.e., fewer) components that represents $\calM\s{b}$ well.
The likelihood of a random sequence $y_{1:\tau} \sim\calM\s{b}$ is given by 
	\begin{eqnarray}
	p(y_{1:\tau}|\calM\s{b}) = \sum_{i=1}^{K\s{b}} \omega_i\s{b} p(y_{1:\tau}|z\s{b}=i,\calM\s{b}) ,
	\label{eqn:mixmodel_base}
	\end{eqnarray}
 where $z\s{b}\sim\mathrm{multinomial}(\omega_1\s{b}, \cdots \omega_{K\s{b}}\s{b})$
 is the hidden variable that indexes the mixture components.  $p(y_{1:\tau}|z=i,\calM\s{b})$ is the likelihood of $y_{1:\tau}$ under the $i$th mixture component, as in \refeqn{eqn:HMM-obs}, and $\omega_i\s{b}$ is the mixture weight for the $i$th component.  
Likewise, the likelihood of the random sequence $y_{1:\tau}\sim\calM\s{r}$ is
	\begin{eqnarray}
	p(y_{1:\tau}|\calM\s{r}) = \sum_{j=1}^{K\s{r}} \omega_j\s{r} p(y_{1:\tau}|z\s{r}=j,\calM\s{r}) ,
	\label{eqn:mixmodel_reduce}
	\end{eqnarray}
where $z\s{r}\sim\mathrm{multinomial}(\omega_1\s{r},\cdots,\omega_{K\s{r}}\s{r})$ is the hidden variable for indexing components in $\calM\s{r}$.  

\ted{At a high level, the VHEM-H3M algorithm estimates the reduced H3M model $\calM\s{r}$ in \refeqn{eqn:mixmodel_reduce}  from \emph{virtual} sequences distributed according to the base H3M model $\calM\s{b}$ in \refeqn{eqn:mixmodel_base}. From this estimation procedure, the VHEM algorithm provides:}
%
\begin{enumerate}
\item a soft clustering of the original $K\s{b}$ components into $K\s{r}$ groups, where cluster membership is encoded in assignment variables that represent the \emph{responsibility} of each reduced mixture component for each base mixture component, i.e., $\hat z_{i,j} = p(z\s{r} = j | z\s{b} = i)$, for $i=1,\dots,K\s{b}$ and $j = 1,\dots,K\s{r}$;
\item novel cluster centers represented by the individual mixture components of the reduced model in \refeqn{eqn:mixmodel_reduce}, i.e., $p(y_{1:\tau}|z\s{r}=j,\calM\s{r}) $ for $j = 1,\dots,K\s{r}$.
\end{enumerate}
%
%
%
\ted{Finally, because we take the expectation over the virtual samples,}
the estimation 
is carried out in an efficient manner that requires only knowledge of the parameters of the base model, without the need of generating actual virtual samples.

\vspace{.2in}
\noindent {\bf{Notation.}}
\begin{table}[t!]
\caption{Notation used in the derivation of the VHEM-H3M algorithm. 
}
\begin{center}
	\resizebox{\columnwidth}{!}{
\begin{tabular}{lll}
\hline
\multicolumn{1}{c}{\emph{variables}}					& \emph{base model (b)}		& \emph{reduced model (r)} \\
index for HMM components						& $i$					& $j$\\
number of HMM components						& $K\s{b}$			& $K\s{r}$ \\
HMM states 									& $\beta$   			& $\rho$   \\
number of HMM states							& $S$				& $S$ \\
HMM state sequence							& $\beta_{1:\tau} = \{\beta_1,\cdots,\beta_{\tau}\}$		& $\rho_{1:\tau} = \{\rho_1,\cdots,\rho_\tau\}$ \\
index for component of GMM						& $m$				& $\ell$ \\
number of Gaussian components					& $M$				& $M$ \\
%
\multicolumn{1}{c}{\emph{models}}		\\			
H3M											& $\calM\s{b}$			& $\calM\s{r}$ \\
HMM component (of H3M)  								& $\calM\s{b}_i$		& $\calM\s{r}_j$ \\
GMM emission									& $\calM\s{b}_{i,\beta}$ 	&$\calM\s{r}_{j,\rho}$ \\
Gaussian component (of GMM)								&$\calM^{(b)}_{i,\beta,m}$& $\calM^{(r)}_{j,\rho,\ell}$ \\
\multicolumn{1}{c}{\emph{parameters}} \\
H3M mixture weights				& $\omega\s{b} = \{\omega\s{b}_i\}$   
							& $\omega\s{r} = \{\omega\s{r}_j\}$ \\  
HMM initial state 				& $\pi^{(b),i} = \{\pi^{(b),i}_{\beta}\}$   
							& $\pi^{(r),j}=\{\pi^{(r),j}_{\rho}\}$ \\  
HMM state transition matrix 		& $A^{(b),i} = [a^{(b),i}_{\beta,\beta'}]$ & $A^{(r),j} = [a^{(r),j}_{\rho,\rho'}]$ \\
GMM emission 					& $\{c^{(b),i}_{\beta,m}, \mu^{(b),i}_{\beta,m}, \Sigma^{(b),i}_{\beta,m}\}_{m=1}^M$ 
							& $\{c^{(r),j}_{\rho,\ell}, \mu^{(r),j}_{\rho,\ell}, \Sigma^{(r),j}_{\rho,\ell}\}_{\ell=1}^M$ \\
\hline
\multicolumn{1}{c}{\emph{probability distributions}}	& \multicolumn{1}{l}{\emph{notation}}	& \multicolumn{1}{l}{\emph{short-hand}}	 \\
HMM state sequence (b) & $p(x_{1:\tau} = \beta_{1:\tau} | z\s{b}=i, \calM\s{b})$	&  $p(\beta_{1:\tau}|\calM\s{b}_{i}) = \pi^{(b),i}_{\beta_{1:\tau}}$   \\ 
HMM state sequence (r) &  $p(x_{1:\tau} = \rho_{1:\tau} | z\s{r}=j, \calM\s{r})$	&  $p(\rho_{1:\tau}|\calM\s{r}_j) = \pi^{(r),j}_{\rho_{1:\tau}}$   \\
HMM observation likelihood (r) & $p(y_{1:\tau}|z\s{r}=j,\calM\s{r})$			& $p(y_{1:\tau}|\calM\s{r}_j)$ \\
GMM emission likelihood (r)  &  $p(y_{t}| x_{t} = \rho,\calM\s{r}_j)$ 			& $p(y_{t}|\calM\s{r}_{j,\rho})$ \\
Gaussian component likelihood (r)  &  $p(y_{t}| \zeta_t=\ell, x_{t} = \rho,\calM\s{r}_j)$ 			& $p(y_{t}|\calM\s{r}_{j,\rho,\ell})$ \\
\multicolumn{1}{c}{\emph{expectations}} & & \\
HMM observation sequence (b) & $\EV_{y_{1:\tau}|z\s{b}=i,\calM\s{b}} [\cdot ]$					& $\EV_{\calM\s{b}_i} [\cdot ]$ \\
GMM emission (b) & $\EV_{y_{t}|x_{t} = \beta,\calM\s{b}_i} [\cdot ]$ 	& $\EV_{\calM\s{b}_{i,\beta}} [\cdot ]$\\
Gaussian component (b) & $\EV_{y_{t}|\zeta_t=m,x_{t} = \beta,\calM\s{b}_i} [\cdot ]$ 	& $\EV_{\calM\s{b}_{i,\beta,m}} [\cdot ]$\\
\hline
\multicolumn{1}{c}{\emph{expected log-likelihood}}		
	& \multicolumn{1}{l}{\emph{lower bound}}			
	& \multicolumn{1}{l}{\emph{variational distribution}}	 \\
$\EV_{\calM\s{b}_i}[\log p(Y_i|\calM\s{r})]$		
	& $\calL_{H3M}^{i}$	
	& $\calP_i(z_i = j) = z_{ij}$	\\
$\EV_{\calM\s{b}_i}[\log p(y_{1:\tau}|\calM\s{r}_j)]$
	& $\calL_{HMM}^{i,j}$
	& $ \calP^{i,j} (\rho_{1:\tau}|\beta_{1:\tau}) =    \phi^{i,j}_{\rho_{1:\tau}|\beta_{1:\tau}}$ \\
	& & \quad$= \phi^{i,j}_{1}(\rho_1|\beta_1) \prod_{t=2}^{\tau} \phi^{i,j}_{t}(\rho_t|\rho_{t-1}, \beta_t)$	\\
$\EV_{\calM\s{b}_{i, \beta}} [\log p(y|\calM\s{r}_{j,\rho})]$ 
	& $\calL_{GMM}^{(i,\beta),(j,\rho)}$
	& $\calP_{\beta,\rho}^{i,j} (\zeta = \ell|m) = \eta^{(i,\beta),(j,\rho)}_{\ell|m}$	\\
\hline
\end{tabular}
}
\end{center}
\label{tab:variables}
\end{table}
We will always use $i$ and $j$ to index the components of the base model $\calM\s{b}$ and the reduced model $\calM\s{r}$, respectively.  
To reduce clutter, we will also use the short-hand notation $\calM\s{b}_i$ and $\calM\s{r}_j$ to denote the $i$th component of $\calM\s{b}$ and the $j$th component of $\calM\s{r}$, respectively.
Hidden states of the HMMs are denoted with $\beta$ for the base model $\calM\s{b}_i$, and with $\rho$ for the reduced model $\calM\s{r}_j$.  

The GMM emission models for each hidden state are denoted as $\calM\s{b}_{i,\beta}$ and $\calM\s{r}_{j,\rho}$.
We will always use $m$ and $\ell$ for indexing the individual Gaussian components of the GMM emissions of the base and reduced models, respectively.  The individual Gaussian components are denoted as $\calM^{(b)}_{i,\beta,m}$ for the base model, and $\calM^{(r)}_{j,\rho,\ell}$ for the reduced model.
Finally, we denote the parameters of $i$th HMM component of the base mixture model as
$\calM\s{b}_i = \{\pi^{(b),i}, A^{(b),i}, \{\{c^{(b),i}_{\beta,m}, \mu^{(b),i}_{\beta,m}, \Sigma^{(b),i}_{\beta,m}\}_{m=1}^{M}\}_{\beta=1}^{S} \}$, and for the $j$th HMM in the reduced mixture as
$\calM\s{r}_j = \{\pi^{(r),j}, A^{(r),j}, \{\{c^{(r),j}_{\rho,\ell}, \mu^{(r),j}_{\rho,\ell}, \Sigma^{(r),j}_{\rho,\ell}\}_{\ell=1}^{M}\}_{\rho=1}^{S} \}$.

When appearing in a probability distribution, the short-hand model notation (e.g., $\calM\s{b}_i$) always implies {\em conditioning} on the model being active.   For example, we will use $p(y_{1:\tau}|\calM\s{b}_i)$ as short-hand for $p(y_{1:\tau}|z\s{b}=i,\calM\s{b})$, or $p(y_{t}|\calM\s{b}_{i,\beta})$ as short-hand for $p(y_{t}|x_t=\beta,z\s{b}=i,\calM\s{b})$.
Furthermore, we will use $\pi^{(b),i}_{\beta_{1:\tau}}$ as short-hand for the probability of the state sequence $\beta_{1:\tau}$ according to the base HMM component $\calM\s{b}_i$, i.e., $p(\beta_{1:\tau}|\calM\s{b}_i)$, and likewise $\calM^{(r),j}_{\rho_{1:\tau}}$ for the reduced HMM component.

Expectations will also use the short-hand model notation to imply conditioning on the model.
In addition, expectations are assumed to be taken with respect to the output variable ($y_{1:\tau}$ or $y_t$), unless otherwise specified.  For example, we will use $\EV_{\calM\s{b}_i} [\cdot ]$  as short-hand for $\EV_{y_{1:\tau}|z\s{b}=i,\calM\s{b}} [\cdot ]$.


Table \ref{tab:variables} summarizes the notation used in the derivation, including the variable names, model parameters, and short-hand notations for probability distributions and expectations.  The bottom of Table \ref{tab:variables} also summarizes the variational lower bound and variational distributions, which will be introduced subsequently.

\subsection{Variational HEM algorithm } \label{subsev:var_lb}

To learn the reduced model in \refeqn{eqn:mixmodel_reduce}, we consider a set of $N$ {\em virtual} samples, distributed according to the base model $\calM\s{b}$ in \refeqn{eqn:mixmodel_base}, such that $N_i = N\omega\s{b}_i$ samples are drawn from the $i$th component.  We denote the set of $N_i$ virtual samples for the $i$th component as $Y_i = \{y\s{i,m}_{1:\tau}\}_{m=1}^{N_i}$, where $y\s{i,m}_{1:\tau}\sim \calM\s{b}_i$, and the entire set of $N$ samples as $Y = \{Y_i\}_{i=1}^{K\s{b}}$.  
Note that, in this formulation, we are not considering virtual samples $\{x\s{i,m}_{1:\tau}, y\s{i,m}_{1:\tau}\}$ for each base component, according to its joint distribution $p(x_{1:\tau},y_{1:\tau}|\calM\s{b}_i)$.  The reason is that the hidden-state space of each base mixture component $\calM\s{b}_i$ may have a different representation (\eg the numbering of the hidden states may be permuted between the components).  This  mismatch will cause problems when the parameters of $\calM\s{r}_j$ are computed from virtual samples of the hidden states of $\{\calM\s{b}_i\}_{i=1}^{K\s{b}}$.  Instead, we treat $X_i=\{x_{1:\tau}\s{i,m}\}_{m=1}^{N_i}$ as ``missing'' information, and estimate them in the E-step. 
The log-likelihood of the virtual samples is
\begin{align}
	\log p(Y|\calM\s{r}) = \sum_{i=1}^{K\s{b}} \log p(Y_i|\calM\s{r}), 
	\label{eqn:LL} 
\end{align}
where, in order to obtain a consistent clustering, we assume the entirety of samples $Y_i$ is assigned to 	the same component of the reduced model \citep{Vasc1998}.  

The original formulation of HEM \citep{Vasc1998} maximizes \refeqn{eqn:LL} with respect to $\calM\s{r}$, and uses the law of large numbers to turn the virtual samples $Y_i$ into an expectation over the base model components $\calM\s{b}_i$.  In this paper,  we will start with a  different objective function to derive the VHEM algorithm. 
To estimate $\calM\s{r}$, we will maximize the average log-likelihood of all possible virtual samples, weighted by their likelihood of being generated by $\calM\s{b}_i$, i.e., the {\em expected} log-likelihood of the virtual samples,
\begin{align}
{\calJ({\calM\s{r}})} = 
	\EV_{\calM\s{b}} \left[ \log p(Y|\calM\s{r}) \right] 
	= \sum_{i=1}^{K\s{b}} \EV_{\calM\s{b}_i} \left[\log p(Y_i|\calM\s{r}) \right],
	\label{eqn:cost_function} 
\end{align}
where the expectation is over the base model components $\calM\s{b}_i$.   
Maximizing \refeqn{eqn:cost_function} will eventually lead to the same estimate as maximizing \refeqn{eqn:LL}, but allows us to strictly preserve the variational lower bound, which would otherwise be ruined when applying the law of large numbers to \refeqn{eqn:LL}.

A general approach to deal with maximum likelihood estimation in the presence of hidden variables (which is the case for H3Ms) is the EM algorithm \citep{Dempster1977}.
In the traditional formulation the EM algorithm is presented as an alternation between an expectation step (E-step) and a maximization step (M-step).
In this work, we take a variational perspective \citep{neal1998view,wainwright2008graphical,csisz1984information}, which views each step as a maximization step.
The variational E-step first obtains a family of lower bounds to the (expected) log-likelihood (i.e., to  (\ref{eqn:cost_function})), indexed by variational parameters, and then optimizes over the variational parameters to find the tightest bound.  
The corresponding M-step then maximizes the lower bound (with the variational parameters fixed) with respect to the model parameters.
One advantage of the variational formulation is that it readily allows for useful extensions to the EM algorithm, such as  replacing a difficult inference in the E-step with a variational approximation. In practice, this is achieved by restricting the maximization in the variational E-step to a smaller domain for which the lower bound is tractable.

\subsubsection{Lower bound to an expected log-likelihood}

Before proceeding with the derivation of VHEM for H3Ms, we first need to derive a lower-bound to an expected log-likelihood term, e.g., \refeqn{eqn:cost_function}.  
In all generality, let $\{O,H\}$ be the observation and hidden variables of a probabilistic model, respectively, where $p(H)$ is the distribution of the hidden variables, $p(O|H)$ is the conditional likelihood of the observations, and $p(O)=\sum_H p(O|H)p(H)$ is the observation likelihood.
We can define a \emph{variational lower bound} to the observation log-likelihood  
\citep{jordan1999introduction,Jaakkola00tutorialon}:
\begin{align}
	\log p(O)  &\geq  \log p(O) - D(\calP(H)|| p(H | O) )    \label{eqn:vlb1}
                                           \\
 &= \sum_H \calP(H)  \log  \frac{p(H) p(O|H)}{ \calP(H)}, 
	\label{eqn:LB-LL}
\end{align}
where $p(H | O)$ is the posterior distribution of $H$ given observation $O$, 
and
$D(p\|q) = \int p(y) \log \frac{p(y)}{q(y)} dy$ is the Kullback-Leibler (KL) divergence between two distributions, $p$ and $q$.
We introduce a variational distribution $\calP(H)$, which approximates the posterior distribution, where $\sum_{H} \calP(H) = 1$ and $\calP(H) \geq 0$.
When the variational distribution equals the true posterior, $\calP(H) = P(H | O)$, then the KL divergence is zero, and hence the lower-bound reaches $\log p(O)$.
When the true posterior cannot be computed, then typically $\calP$ is restricted to some set of approximate posterior distributions $\cal Q$ that are tractable, and the best lower-bound is obtained by maximizing over $\calP\in{\cal Q}$,
\begin{align}
	\log p(O)  &\geq  \max_{\calP\in {\cal Q}} \sum_H \calP(H)  \log  \frac{p(H) p(O|H)}{ \calP(H)}. 
	\label{eqn:LB-maxLL}
\end{align}

Using the lower bound in 	\refeqn{eqn:LB-maxLL}, we can now derive a lower bound to an expected log-likelihood expression.  Let $\EV_b[\cdot]$ be the expectation with respect to $O$ with some distribution $p_b(O)$.  Since $p_b(O)$ is non-negative, taking the expectation on both sides of $\refeqn{eqn:LB-maxLL}$ yields,
	\begin{align}
	\EV_b\left[\log p(O)\right]  &\geq  \EV_b\left[\max_{\calP\in Q} \sum_H \calP(H)  \log  \frac{p(H) p(O|H)}{ \calP(H)} \right]
	\label{eqn:LB-ELL1} \\
	&\geq  \max_{\calP\in Q}  \EV_b\left[\sum_H \calP(H)  \log  \frac{p(H) p(O|H)}{ \calP(H)} \right] 
	\label{eqn:LB-ELL2} \\
	&= \max_{\calP\in Q}  \sum_H \calP(H)  \left\{\log \frac{p(H)}{ \calP(H)}  + \EV_b\left[\log p(O|H) \right] \right\},
	\label{eqn:LB-ELL}
\end{align}
where \refeqn{eqn:LB-ELL2} follows from Jensen's inequality (i.e., $f(\EV[x]) \leq \EV[f(x)]$ when $f$ is convex), and the convexity of the $\max$ function.  Hence, \refeqn{eqn:LB-ELL} is a variational lower bound on the expected log-likelihood, which depends on 
the family of variational distributions $Q$.

\comments{
\tn{This can probably be cut, although it is interesting in a broader context.}
\refeqn{eqn:LB-ELL} can be rewritten in terms of KL divergence and expectations,
	\begin{align}
	\EV_b\left[\log p(O)\right]   &\geq \max_{\calP\in Q} \left[ \sum_H \calP(H) \EV_b\left[\log p(O|H) \right]  \right]
	- D(\calP(H)||p(H)), \\
	&\geq \max_{\calP\in Q}  \EV_{\calP(H)} \left[\EV_b\left[\log p(O|H) \right]  \right]
	- D(\calP(H)||p(H)),
	\end{align}
where $\EV_{\calP(H)}$ is the expectation wrt $\calP(H)$.  This gives an interesting intuition.  
The lower bound is composed of two terms: 1) an expectation of the conditional log-likelihood $\log p(O|H)$, over $O$ and marginalized over the approximate distribution $\calP(H)$; 2) a divergence term that encourages $\calP(H)$ to be similar to $p(H)$.
}

\subsubsection{Variational lower bound}

We now derive a lower bound to the expected log-likelihood cost function in \refeqn{eqn:cost_function}.  The derivation will proceed by successively applying the lower bound from \refeqn{eqn:LB-ELL} to each expected log-likelihood term that arises. This will result in a set of nested lower bounds.  We first define the following three lower bounds:
	\begin{align}
	\EV_{\calM\s{b}_i}[\log p(Y_i|\calM\s{r})] &\geq \calL_{H3M}^{i}, 
	\label{eqn:LB-H3M}
	\\
	\EV_{\calM\s{b}_i}[\log p(y_{1:\tau}|\calM\s{r}_j)] &\geq \calL_{HMM}^{i,j} ,
	\label{eqn:LB-HMM}
	\\
	\EV_{\calM\s{b}_{i, \beta}} [\log p(y|\calM\s{r}_{j,\rho})] &\geq \calL_{GMM}^{(i,\beta),(j,\rho)}.
	\label{eqn:LB-GMM}
	\end{align}
The first lower bound, $\calL_{H3M}^{i}$, is on the expected log-likelihood of an H3M $\calM\s{r}$ with respect to an HMM $\calM\s{b}_i$.
The second lower bound, $\calL_{HMM}^{i,j}$, is on the expected log-likelihood of an HMM $\calM\s{r}_j$, averaged 
over observation sequences from a \emph{different} HMM $\calM\s{b}_i$.  Although the data log-likelihood $\log p(y_{1:\tau}|\calM\s{r}_j)$ can be computed exactly using the forward algorithm \citep{Rabiner93}, calculating its expectation is not analytically tractable since 
an observation sequence $y_{1:\tau}$ from a HMM $\calM\s{r}_j$ is 
essentially an observation from a mixture model.\footnote{For an observation sequence of length $\tau$, an HMM with $S$ states can be considered as a mixture model with  $O(S^\tau)$ components.}
The third lower bound, $\calL_{GMM}^{(i,\beta),(j,\rho)}$, is on the expected log-likelihood
of a GMM emission density $\calM\s{r}_{j,\rho}$ with respect to another GMM $\calM\s{b}_{i, \beta}$.  This lower bound does not depend on time, as we have assumed that the emission densities are time-invariant.

Looking at an individual term in \refeqn{eqn:cost_function}, $p(Y_i|\calM\s{r})$ is the likelihood under a mixture of HMMs, as in \refeqn{eqn:mixmodel_reduce}, where
the observation variable is $Y_i$ and the hidden variable is $z_i$ (the assignment of $Y_i$ to a component of $\calM\s{r}$).  Hence, introducing the variational distribution $\calP_i(z_i)$ and applying  \refeqn{eqn:LB-ELL}, we have
	\begin{align}
	\nonumber
	\lefteqn{ \EV_{\calM\s{b}_i} \left[\log p(Y_i|\calM\s{r}) \right] } \\
	 &\geq \max_{\calP_i} \sum_{j} \calP_i(z_i=j) \left\{ \log \frac{p(z_i=j|\calM\s{r})}{\calP_i(z_i=j)} 
	 +  \EV_{\calM\s{b}_i}[\log p(Y_i|\calM\s{r}_j)]  \right\} 
	 \\
	 &= \max_{\calP_i} \sum_{j} \calP_i(z_i=j) \left\{ \log \frac{p(z_i=j|\calM\s{r})}{\calP_i(z_i=j)} 
	 +  \EV_{\calM\s{b}_i}[\log p(y_{1:\tau}|\calM\s{r}_j)^{N_i}]  \right\}
	 \label{eqn:iidY}
	 \\
	 &= \max_{\calP_i} \sum_{j} \calP_i(z_i=j) \left\{ \log \frac{p(z_i=j|\calM\s{r})}{\calP_i(z_i=j)} 
	 +  N_i\EV_{\calM\s{b}_i}[\log p(y_{1:\tau}|\calM\s{r}_j)]  \right\}
	\label{eqn:LB-H3M-1}
	,
	\end{align}
where in \refeqn{eqn:iidY} we use the fact that $Y_i$ is a set of $N_i$ i.i.d. samples.  In \refeqn{eqn:LB-H3M-1}, $\log p(y_{1:\tau}|\calM\s{r}_j)$ is the observation log-likelihood of an HMM, which is essentially a mixture distribution, and hence its expectation cannot be calculated directly.  Instead, we use the lower bound $\calL_{HMM}^{i,j}$ defined in \refeqn{eqn:LB-HMM}, yielding
	\begin{align}
	 \EV_{\calM\s{b}_i} \left[\log p(Y_i|\calM\s{r}) \right] \geq 
	\max_{\calP_i} \sum_{j} \calP_i(z_i=j) \left\{ \log \frac{p(z_i=j|\calM\s{r})}{\calP_i(z_i=j)} 
	 +  N_i \calL_{HMM}^{i,j} \right\} \triangleq \calL_{H3M}^i.
	 \label{eqn:LB-H3M-defn}
	 \end{align}

Next, we calculate the lower bound $\calL_{HMM}^{i,j}$. Starting with \refeqn{eqn:LB-HMM},  we first rewrite the expectation $\EV_{\calM\s{b}_i}$ to explicitly marginalize over the HMM state sequence $\beta_{1:\tau}$ from $\calM\s{b}_i$, 
	\begin{align}
	\EV_{\calM\s{b}_i}[\log p(y_{1:\tau}|\calM\s{r}_j)] &= 
	\EV_{\beta_{1:\tau}|\calM\s{b}_i} \left[\EV_{y_{1:\tau}|\beta_{1:\tau},\calM\s{b}_{i}}[\log p(y_{1:\tau}|\calM\s{r}_j)]\right]
	\\
	&= \sum_{\beta_{1:\tau}} \pi_{\beta_{1:\tau}}^{(b),i} \EV_{y_{1:\tau}|\beta_{1:\tau},\calM\s{b}_{i}}[\log p(y_{1:\tau}|\calM\s{r}_j)].
	\label{eqn:LB-HMM1}
	\end{align}
For the HMM likelihood $p(y_{1:\tau}|\calM\s{r}_j)$, given by \refeqn{eqn:HMM-obs}, the observation variable is $y_{1:\tau}$ and the hidden variable is the state sequence $\rho_{1:\tau}$.   We therefore introduce a variational distribution $\calP^{i,j}(\rho_{1:\tau}|\beta_{1:\tau})$ on the state sequence $\rho_{1:\tau}$, which depends on a particular sequence $\beta_{1:\tau}$ from $\calM\s{b}_i$.  Applying \refeqn{eqn:LB-ELL} to \refeqn{eqn:LB-HMM1}, we have
	\begin{align}
	\nonumber
	\lefteqn{\EV_{\calM\s{b}_i}[\log p(y_{1:\tau}|\calM\s{r}_j)]}
	\\
	&\geq
	\sum_{\beta_{1:\tau}} \pi_{\beta_{1:\tau}}^{(b),i}
	\max_{\calP^{i,j}} \sum_{\rho_{1:\tau}} 
	\calP^{i,j}(\rho_{1:\tau}|\beta_{1:\tau}) \left\{
	\log \frac{p(\rho_{1:\tau}|\calM\s{r}_j)}{\calP^{i,j}(\rho_{1:\tau}|\beta_{1:\tau}) }
	+ \EV_{y_{1:\tau}|\beta_{1:\tau},\calM\s{b}_{i}} [\log p(y_{1:\tau}|\rho_{1:\tau},\calM\s{r}_j)]
	\right\}
	\\
	&=
	\sum_{\beta_{1:\tau}} \pi_{\beta_{1:\tau}}^{(b),i}
	\max_{\calP^{i,j}} \sum_{\rho_{1:\tau}} 
	\calP^{i,j}(\rho_{1:\tau}|\beta_{1:\tau}) \left\{
	\log \frac{p(\rho_{1:\tau}|\calM\s{r}_j)}{\calP^{i,j}(\rho_{1:\tau}|\beta_{1:\tau}) }
	+ \sum_t \EV_{\calM\s{b}_{i,\beta_{t}}} [\log p(y_t|\calM\s{r}_{j,\rho_t})]
	\right\}
	\label{eqn:use_Markov}
	\\
	&\geq 
	\sum_{\beta_{1:\tau}} \pi_{\beta_{1:\tau}}^{(b),i}
	\max_{\calP^{i,j}} \sum_{\rho_{1:\tau}} 
	\calP^{i,j}(\rho_{1:\tau}|\beta_{1:\tau}) \left\{
	\log \frac{p(\rho_{1:\tau}|\calM\s{r}_j)}{\calP^{i,j}(\rho_{1:\tau}|\beta_{1:\tau}) }
	+ \sum_t \calL_{GMM}^{(i,\beta_t),(j,\rho_t)}
	\right\}
	\triangleq \calL_{HMM}^{i,j},
	 \label{eqn:LB-HMM-defn}
	\end{align}
where in \refeqn{eqn:use_Markov} 
we use the conditional independence of the observation sequence given the state sequence, 
and in \refeqn{eqn:LB-HMM-defn} we use the lower bound, defined in \refeqn{eqn:LB-GMM}, on each expectation.

Finally, we derive the lower bound $\calL_{GMM}^{(i,\beta),(j,\rho)}$ for \refeqn{eqn:LB-GMM}.  First, we rewrite the expectation with respect to $\calM\s{b}_{i,\beta}$ to explicitly marginalize out the GMM hidden assignment variable $\zeta$,
	\begin{align}
	\EV_{\calM\s{b}_{i,\beta}} [\log p(y|\calM\s{r}_{j,\rho})]
	&=
	\EV_{\zeta|\calM\s{b}_{i,\beta}} \left[ \EV_{\calM\s{b}_{i,\beta,\zeta}} [\log p(y|\calM\s{r}_{j,\rho})]\right]
	\\
	&=
	\sum_{m=1}^M c^{(b),i}_{\beta,m} 
	\EV_{\calM\s{b}_{i,\beta,m}} \left[\log p(y|\calM\s{r}_{j,\rho})\right].
	\end{align}
Note that $p(y|\calM\s{r}_{j,\rho})$ is a GMM emission distribution as in \refeqn{eqn:GMM}. Hence, the observation variable is $y$, and the hidden variable is $\zeta$.  Therefore, we introduce the variational distribution $\calP^{i,j}_{\beta,\rho}(\zeta|m)$, which is conditioned on the observation $y$ arising from the $m$th component in $\calM\s{b}_{i,\beta}$, and apply \refeqn{eqn:LB-ELL}, 
	\begin{align}
	\nonumber
	\lefteqn{
	\EV_{\calM\s{b}_{i,\beta}} [\log p(y|\calM\s{r}_{j,\rho})] 
	}\\
	&\geq
	\sum_{m=1}^M c^{(b),i}_{\beta,m} 
	\max_{\calP^{i,j}_{\beta,\rho}} \sum_{\zeta=1}^M \calP^{i,j}_{\beta,\rho}(\zeta|m)\left\{
	\log \frac{p(\zeta|\calM\s{r}_{j,\rho})}{\calP^{i,j}_{\beta,\rho}(\zeta|m)}
	+
	\EV_{\calM\s{b}_{i,\beta,m}}[\log p(y|\calM\s{r}_{j,\rho,\zeta})]
	\right\}
	\triangleq \calL_{GMM}^{(i,\beta),(j,\rho)},
	\end{align}
where $\EV_{\calM\s{b}_{i,\beta,m}}[\log p(y|\calM\s{r}_{j,\rho,\ell})]$ is the expected log-likelihood of the Gaussian distribution $\calM\s{r}_{j,\rho,\ell}$ with respect to the Gaussian $\calM\s{b}_{i,\beta,m}$, which has a closed-form solution (see Section \ref{text:VD}).

In summary, we have derived a variational lower bound to the expected log-likelihood of the virtual samples in \refeqn{eqn:cost_function},
\begin{align}
\label{eqn:LB-cost-summary}
{\calJ({\calM\s{r}})} =
	\EV_{\calM\s{b}} \left[ \log p(Y|\calM\s{r}) \right]  \geq
	 \sum_{i=1}^{K\s{b}}  \calL_{H3M}^i,
\end{align}
which is composed of three nested lower bounds, corresponding to different model elements (the H3M, the component HMMs, and the emission GMMs),
\begin{align}
	\label{eqn:LB-H3M-summary}
	\calL_{H3M}^i &= 
	\max_{\calP_i} \sum_{j} \calP_i(z_i=j) \left\{ \log \frac{p(z_i=j|\calM\s{r})}{\calP_i(z_i=j)} 
	 +  N_i \calL_{HMM}^{i,j} \right\} ,
	\\
	\label{eqn:LB-HMM-summary}
	 \calL_{HMM}^{i,j} &= 
	\sum_{\beta_{1:\tau}} \pi_{\beta_{1:\tau}}^{(b),i}
	\max_{\calP^{i,j}} \sum_{\rho_{1:\tau}} 
	\calP^{i,j}(\rho_{1:\tau}|\beta_{1:\tau}) \left\{
	\log \frac{p(\rho_{1:\tau}|\calM\s{r}_j)}{\calP^{i,j}(\rho_{1:\tau}|\beta_{1:\tau}) }
	+ \sum_t \calL_{GMM}^{(i,\beta_t),(j,\rho_t)}
	\right\}  ,
	\\
	\label{eqn:LB-GMM-summary}
	\calL_{GMM}^{(i,\beta),(j,\rho)} &= 
	\sum_{m=1}^M c^{(b),i}_{\beta,m} 
	\max_{\calP^{i,j}_{\beta,\rho}} \sum_{\zeta=1}^M \calP^{i,j}_{\beta,\rho}(\zeta|m)\left\{
	\log \frac{p(\zeta|\calM\s{r}_{j,\rho})}{\calP^{i,j}_{\beta,\rho}(\zeta|m)}
	+
	\EV_{\calM\s{b}_{i,\beta,m}}[\log p(y|\calM\s{r}_{j,\rho,\zeta})]
	\right\},
\end{align}
where $\calP_i(z_i)$, $\calP^{i,j}(\rho_{1:\tau}|\beta_{1:\tau})$, and $\calP^{i,j}_{\beta,\rho}(\zeta|m)$ are the corresponding variational distributions.  Finally, the variational HEM algorithm for HMMs consists of two alternating steps:
\begin{itemize}
\item (variational E-step) given $\calM\s{r}$, calculate the variational distributions $\calP^{i,j}_{\beta,\rho}(\zeta|m)$, $\calP^{i,j}(\rho_{1:\tau}|\beta_{1:\tau})$, and $\calP_i(z_i)$ for the lower bounds in \refeqn{eqn:LB-GMM-summary}, \refeqn{eqn:LB-HMM-summary}, and \refeqn{eqn:LB-H3M-summary};
\item (M-step) update the model parameters via ${\calM\s{r}}^*  = \argmax_{\calM\s{r}} \sum_{i=1}^{K\s{b}}  \calL_{H3M}^i$.
\end{itemize}
In the following subsections, we derive the E- and M-steps of the algorithm.


\subsection{Variational E-step } \label{subsev:e:new}

The variational E-step consists of finding the variational distributions that maximize the lower bounds in  \refeqn{eqn:LB-GMM-summary}, \refeqn{eqn:LB-HMM-summary}, and \refeqn{eqn:LB-H3M-summary}.  
In particular, given the nesting of the lower bounds, we proceed by first maximizing the GMM lower bound $\calL_{GMM}^{(i,\beta),(j,\rho)}$ for each pair of emission GMMs in the base and reduced models.
Next, the HMM lower bound $\calL_{HMM}^{i,j}$ is maximized for each pair of HMMs in the base and reduced models, 
followed by maximizing the H3M lower bound $\calL_{H3M}^i$ for each base HMM.  
Finally, a set of summary statistics are calculated, which will be used in the M-step.

\subsubsection{Variational distributions}
\label{text:VD}

We first consider the forms of the three variational distributions, as well as the optimal parameters to maximize the corresponding lower bounds.

{\bf GMM}:
For the GMM lower bound $\calL_{GMM}^{(i,\beta),(j,\rho)}$, we assume each
variational distribution has the form \citep{hershey2008variational}
	\begin{align}
	\calP^{i,j}_{\beta,\rho}(\zeta=l | m) = \eta^{(i,\beta),(j,\rho)}_{\ell|m},
	\end{align}
where $\sum_{\ell=1}^{M} \eta^{(i,\beta),(j,\rho)}_{\ell|m} =1$, and $\eta^{(i,\beta),(j,\rho)}_{\ell|m} \geq 0$, $\forall \ell$.
Intuitively,  $\boldsymbol\eta^{(i,\beta),(j,\rho)}$ is the responsibility matrix between each pair of Gaussian components in the GMMs $\calM\s{b}_{i,\beta}$ and $\calM\s{r}_{j,\rho}$, 
where $\eta^{(i,\beta),(j,\rho)}_{\ell|m}$ represents the probability that an observation from component $m$ of $\calM\s{b}_{i,\beta}$ corresponds to component $\ell$ of $\calM\s{r}_{j,\rho}$. 


Substituting into \refeqn{eqn:LB-GMM-summary}, we have
	\begin{align}	
	\calL_{GMM}^{(i,\beta),(j,\rho)} &= 
	\sum_{m=1}^M c^{(b),i}_{\beta,m} 
	\max_{\eta^{(i,\beta),(j,\rho)}_{\ell|m}} \sum_{\ell=1}^M \eta^{(i,\beta),(j,\rho)}_{\ell|m} \left\{
	\log \frac{c^{(r),j}_{\rho,\ell}}{\eta^{(i,\beta),(j,\rho)}_{\ell|m}}
	+
	\EV_{\calM\s{b}_{i,\beta,m}}[\log p(y|\calM\s{r}_{j,\rho,\ell})]
	\right\}.
	\label{eqn:LB-GMM-spec}
	\end{align}
The maximizing variational parameters are obtained as (see Appendix C.2)
	\begin{align}
	\hat{\eta}^{(i,\beta),(j,\rho)}_{\ell|m} = \frac{c^{(r),j}_{\rho,\ell} \exp\left\{ \EV_{\calM\s{b}_{i,\beta,m}}[\log p(y|\calM\s{r}_{j,\rho,\ell})]\right\}}{ \sum_{\ell'}c^{(r),j}_{\rho,\ell'}\exp\left\{\EV_{\calM\s{b}_{i,\beta,m}}[\log p(y|\calM\s{r}_{j,\rho,\ell'})]\right\}},
	\end{align}
where the expected log-likelihood of a Gaussian $\calM\s{r}_{j,\rho,\ell}$ with respect to another Gaussian $\calM\s{b}_{i,\beta,m}$ is computable in closed-form \citep{penny2000notes}, 
	\begin{align}
	\begin{split}
	\EV_{\calM\s{b}_{i,\beta,m}}[\log p(y|\calM\s{r}_{j,\rho,\ell})]
	&= 
	-\frac{d}{2} \log 2\pi - \frac{1}{2} \log \left|\Sigma^{(r),j}_{\rho,\ell}\right|  -\frac{1}{2} \mbox{tr} \left( ({\Sigma^{(r),j}_{\rho,\ell}})^{-1} \Sigma^{(b),i}_{\beta,m}\right) 
	\\
	&\quad
 - \frac{1}{2} (\mu^{(r),j}_{\rho,\ell} - \mu^{(b),i}_{\beta,m})^T({\Sigma^{(r),j}_{\rho,\ell}})^{-1} (\mu^{(r),j}_{\rho,\ell} - \mu^{(b),i}_{\beta,m}) .
 \end{split}
	\end{align}
	

{\bf HMM}:
For the HMM lower bound $\calL_{HMM}^{i,j}$, we assume each variational distribution takes the form of a Markov chain,
	\begin{align}
	\calP^{i,j}(\rho_{1:\tau}|\beta_{1:\tau}) 
	= \phi^{i,j}(\rho_{1:\tau}|\beta_{1:\tau})
	 =    \phi^{i,j}_{1}(\rho_1|\beta_1) \prod_{t=2}^{\tau} \phi^{i,j}_{t}(\rho_t|\rho_{t-1}, \beta_t),
	 \label{eqn:VHMM}
	\end{align}
where $\sum_{\rho_1 = 1}^{S}  \phi^{i,j}_{1}(\rho_1|\beta_1) = 1,$ 
and $\sum_{\rho_t = 1}^{S}   \phi^{i,j}_{t}(\rho_t|\rho_{t-1},\beta_t)=1$, and all the factors are non-negative. 
%
%
The variational distribution $\calP^{i,j}(\rho_{1:\tau}|\beta_{1:\tau})$ represents the probability of the state sequence  $\rho_{1:\tau}$ in HMM $\calM\s{r}_j$, when $\calM\s{r}_j$ is used to explain the {\em observation} sequence generated by $\calM\s{b}_i$ that evolved through state sequence $\beta_{1:\tau}$.

Substituting into \refeqn{eqn:LB-HMM-summary}, we have
	\begin{align}	
	 \calL_{HMM}^{i,j} &= 
	\sum_{\beta_{1:\tau}} \pi_{\beta_{1:\tau}}^{(b),i}
	\max_{\phi^{i,j}} \sum_{\rho_{1:\tau}}
	\phi^{i,j}(\rho_{1:\tau}|\beta_{1:\tau}) \left\{
	\log \frac{\pi^{(r),j}_{\rho_{1:\tau}}}{\phi^{i,j}(\rho_{1:\tau}|\beta_{1:\tau}) }
	+ \sum_t \calL_{GMM}^{(i,\beta_t),(j,\rho_t)}
	\right\}.
\label{eqn:Lhmm}
	\end{align}
The maximization with respect to $\phi^{i,j}_{t}(\rho_t|\rho_{t-1},\beta_t)$ and $\phi^{i,j}_{1}(\rho_1|\beta_1)$ is carried out independently for each pair $(i,j)$, and follows \citep{hershey2008variational}. This is further detailed in Appendix A.  By separating terms and breaking up the summation over $\beta_{1:\tau}$ and $\rho_{1:\tau}$, the optimal $\hat{\phi}^{i,j}_{t}(\rho_t|\rho_{t-1},\beta_t)$ and $\hat{\phi}^{i,j}_{1}(\rho_1|\beta_1)$ can be obtained using an efficient recursive iteration (similar to the forward algorithm).


{\bf H3M}:
For the H3M lower bound $\calL^i_{H3M}$, we assume variational distributions of the form $\calP_i(z_i=j) = z_{ij}$, 
where $\sum_{j = 1}^{K\s{r}} z_{ij} = 1$, and $z_{ij} \geq 0$. 
Substituting into \refeqn{eqn:LB-H3M-summary}, we have
	\begin{align}
	\label{eqn:LB-H3M-spec-1}
	\calL_{H3M}^i &= 
	\max_{z_{ij}} \sum_{j} z_{ij} \left\{ \log \frac{\omega\s{r}_j}{z_{ij}} 
	 +  N_i \calL_{HMM}^{i,j} \right\} .
	\end{align}
The maximizing variational parameters of \refeqn{eqn:LB-H3M-spec-1} are obtained in Appendix C.2,
	\begin{align}
	\hat{z}_{ij} = \frac{\omega\s{r}_j \exp (N_i \calL_{HMM}^{i,j})}{\sum_{j'} \omega\s{r}_{j'} \exp(N_i \calL_{HMM}^{i,j'})}
	 \label{eqn:zijhat}
	 .
	\end{align}
Note that in the standard HEM algorithm derived in \citep{Vasc1998,Chan2010cvpr}, the assignment probabilities $z_{ij}$ are based on the expected log-likelihoods of the components, (e.g., $\EV_{\calM\s{b}_i}[\log p(y_{1:\tau}|\calM\s{r}_j)]$ for H3Ms).
For the variational HEM algorithm, these expectations are now replaced with their lower bounds (in our case, $\calL_{HMM}^{i,j}$).

\subsubsection{Lower bound}

Substituting the optimal variational distributions into \refeqn{eqn:LB-GMM-spec}, \refeqn{eqn:Lhmm}, and \refeqn{eqn:LB-H3M-spec-1} gives the lower bounds,
	\begin{align}
	\calL_{H3M}^i &= 
	\sum_{j} \hat{z}_{ij} \left\{ \log \frac{\omega\s{r}_j}{\hat{z}_{ij}} 
	 +  N_i \calL_{HMM}^{i,j} \right\} ,
	 \label{eqn:LB-H3M-final}
	\\
	 \calL_{HMM}^{i,j} &= 
	\sum_{\beta_{1:\tau}} \pi_{\beta_{1:\tau}}^{(b),i}
	\sum_{\rho_{1:\tau}} 
	\hat{\phi}^{i,j}(\rho_{1:\tau}|\beta_{1:\tau}) \left\{
	\log \frac{\pi^{(r),j}_{\rho_{1:\tau}}}{\hat{\phi}^{i,j}(\rho_{1:\tau}|\beta_{1:\tau}) }
	+ \sum_t \calL_{GMM}^{(i,\beta_t),(j,\rho_t)}
	\right\},
	 \label{eqn:LB-HMM-final}
	\\
	\calL_{GMM}^{(i,\beta),(j,\rho)} &= 
	\sum_{m=1}^M c^{(b),i}_{\beta,m} 
	\sum_{\ell=1}^M \hat{\eta}^{(i,\beta),(j,\rho)}_{\ell|m} \left\{
	\log \frac{c^{(r),j}_{\rho,\ell}}{\hat{\eta}^{(i,\beta),(j,\rho)}_{\ell|m}}
	+
	\EV_{\calM\s{b}_{i,\beta,m}}[\log p(y|\calM\s{r}_{j,\rho,\ell})]
	\right\}.
	 \label{eqn:LB-GMM-final}
	\end{align}
The lower bound $\calL_{HMM}^{i,j}$ requires summing over all sequences $\beta_{1:\tau}$ and $\rho_{1:\tau}$.  This summation can be computed efficiently along with $\hat{\phi}^{i,j}_{t}(\rho_t|\rho_{t-1},\beta_t)$ and $\hat{\phi}^{i,j}_{1}(\rho_1|\beta_1)$ using a recursive algorithm from \citep{hershey2008variational}. This is described in Appendix A.

\subsubsection{Summary Statistics}

After calculating the optimal variational distributions, we calculate the following summary statistics, which are necessary for the M-step:
	\begin{align}
\nu_1^{i,j}(\rho_1,\beta_1) &=  \is{b}{i}{{\beta_1}}{{}} \,\hat\phi^{i,j}_{1}(\rho_1|\beta_1), 
	\label{eq:sum_stat1}
	\\
\xi_{t}^{i,j}(\rho_{t-1},\rho_{t},\beta_t) &= \left( \sum_{\beta_{t-1}=1}^{S}  \nu_{t-1}^{i,j}(\rho_{t-1},\beta_{t-1})  \, a^{(b),i}_{\beta_{t-1},\beta_t} \right)\, \hat\phi^{i,j}_{t}(\rho_t|\rho_{t-1},\beta_t) \mbox{, for } t = 2,\dots,\tau ,
	\label{eq:sum_stat2} 
	\\
\nu_t^{i,j}(\rho_t,\beta_t) &=    \sum_{\rho_{t-1}=1}^{S} \xi_t^{i,j}(\rho_{t-1},\rho_t,\beta_t)
\mbox{, for } t = 2,\dots,\tau ,
	\label{eq:sum_stat3}
	\end{align}
and the aggregate statistics 
	\begin{align}
	\hat\nu_1^{i,j}(\rho) &= \sum_{\beta=1}^{S} \nu_1^{i,j}(\rho,\beta) ,
	\label{eq:agg_stat1}
	 \\
	\hat\nu^{i,j}(\rho,\beta) &= \sum_{t=1}^{\tau} \nu_t^{i,j}(\rho,\beta) \label{eq:agg_stat2}, \\
	\hat\xi^{i,j}(\rho,\rho') &= \sum_{t=2}^{\tau} \sum_{\beta=1}^{S} \xi^{i,j}_t(\rho,\rho',\beta) \label{eq:agg_stat3}
	.
	\end{align}
The statistic $\hat\nu_1^{i,j}(\rho)$ is the expected number of times that the HMM $\calM\s{r}_j$ starts from state $\rho$, when modeling sequences generated by $\calM\s{b}_i$.
The quantity $\hat\nu^{i,j}(\rho,\beta)$ is 
the expected number of times that the HMM $\calM\s{r}_j$ is in state $\rho$ when the HMM $\calM\s{b}_i$ is in state $\beta$, when both HMMs are modeling sequences generated by $\calM\s{b}_i$.
Similarly, the quantity  $\hat\xi^{i,j}(\rho,\rho')$ is the expected number of transitions from state $\rho$ to state $\rho'$  of the HMM $\calM\s{r}_j$, when modeling sequences generated by $\calM\s{b}_i$.

\subsection{M-step}\label{subsev:m}

In the M-step, the lower bound in \refeqn{eqn:LB-cost-summary} is maximized with respect to the parameters $\calM\s{r}$,
	\begin{align}
	{\calM\s{r}}^*  = \argmax_{\calM\s{r}} \sum_{i=1}^{K\s{b}}  \calL_{H3M}^i .
	\end{align}
The derivation of the maximization is presented in Appendix B.
Each mixture component of $\calM\s{r}$ is updated independently according to
	\begin{align}
 {\omega\s{r}_j}^* &= \frac{\sum_{i=1}^{K\s{b}} \hat z_{i,j}} {K\s{b}} ,
 \label{eqn:M-step-omega}
 \\
	 {\is{r}{j}{\rho}{{}}}^* &= \frac{\sum_{i=1}^{K\s{b}} \hat z_{i,j} \omega\s{b}_i \hat\nu_1^{i,j}(\rho)}{\sum_{\rho'=1}^{S} \sum_{i=1}^{K\s{b}} \hat z_{i,j} \omega\s{b}_i \hat\nu_1^{i,j}(\rho'))} \mbox{,}
	 \quad
{a^{(r),j}_{\rho,\rho'}}^* =  \frac{\sum_{i=1}^{K\s{b}} \hat z_{i,j} \omega\s{b}_i \hat\xi^{i,j}(\rho,\rho')}{\sum_{\sigma=1}^{S} \sum_{i=1}^{K\s{b}} \hat z_{i,j} \omega\s{b}_i \hat\xi^{i,j}(\rho,\sigma) } ,
 \label{eqn:M-step-pi-a}
\\
{c^{(r),j}_{\rho,\ell} }^*    &=  \frac{\Omega_{j,\rho}\left(\hat \eta^{(i,\beta),(j,\rho)}_{\ell|m} \right)}{\sum_{\ell'=1}^{M} \Omega_{j,\rho}\left(\hat \eta^{(i,\beta),(j,\rho)}_{\ell'|m}\right) }\mbox{,}
\quad\quad
	{\mu^{(r),j}_{\rho,\ell} }^*    =  \frac{\Omega_{j,\rho}\left( \eta^{(i,\beta),(j,\rho)}_{\ell|m} \,\,\,\mu^{(b),i}_{\beta,m}\right)}{\Omega_{j,\rho}\left(\hat \eta^{(i,\beta),(j,\rho)}_{\ell|m} \right)} ,
	\label{eq:updateTXT}
	\\
	{\Sigma^{(r),j}_{\rho,\ell}}^* &= \frac{\Omega_{j,\rho}\left(\hat \eta^{(i,\beta),(j,\rho)}_{\ell|m}\left[ \Sigma^{(b),i}_{\beta,m}  + (\mu^{(b),i}_{\beta,m}-\mu^{(r),j}_{\rho,\ell}) (\mu^{(b),i}_{\beta,m}-\mu^{(r),j}_{\rho,\ell})^T \right]\right)}{\Omega_{j,\rho}\left(\hat \eta^{(i,\beta),(j,\rho)}_{\ell|m}\right)}
	\label{eq:updateSigmaTXT},
\end{align}
where $\Omega_{j,\rho}(\cdot)$ is the weighted sum operator over all base models, HMM states, and GMM components (i.e., over all tuples $(i,\beta,m)$),
	\begin{eqnarray}
\Omega_{j,\rho}( f(i,\beta,m) ) = \sum_{i=1}^{K\s{b}} \hat z_{i,j} \omega\s{b}_i \sum_{\beta = 1}^{S} \hat\nu^{i,j}(\rho,\beta) \,\sum_{m = 1}^{M}  c^{(b),i}_{\beta,m} \, f(i,\beta,m).
	\label{eqn:weightedsum}
	\end{eqnarray}
Note that the covariance matrices of the reduced models in \refeqn{eq:updateSigmaTXT} include an additional outer-product term, which acts to regularize the covariances of the base models.
This regularization effect derives from the E-step, which averages all possible observations from the base model.


\section{Applications and related work}\label{sec:dis}

In the previous section, we derived the VHEM-H3M algorithm to cluster HMMs.
We now discuss various applications of the algorithm (Section \ref{sec:applications}), and then present some literature that is related to HMM clustering (Section \ref{sec:related}).

\subsection{Applications of the VHEM-H3M algorithm}\label{sec:applications}

The proposed VHEM-H3M algorithm clusters HMMs \emph{directly} through the distributions they represent, and learns \emph{novel} HMM cluster centers 
that compactly represent the structure of each cluster. 

An application of the VHEM-H3M algorithm is in \emph{hierarchical clustering} of HMMs.
In particular, the VHEM-H3M algorithm is used recursively on the 
HMM cluster centers, to produce a bottom-up hierarchy of the input HMMs. 
Since the cluster centers condense the structure of the clusters they represent, the VHEM-H3M algorithm can implicitly leverage rich information on the underlying structure of the clusters, which is expected to impact positively the quality of the resulting hierarchical clustering.


Another application of VHEM is for efficient estimation of  H3Ms from 
data, 
by using a \emph{hierarchical estimation procedure} 
to break the learning problem into smaller pieces. 
First, a data set is split into small (non-overlapping) portions and intermediate HMMs 
 are learned for each portion, via standard EM. Then, the final model is estimated from the intermediate models using the VHEM-H3M algorithm. 
Because VHEM and standard EM 
are based on similar maximum-likelihood principles,
it drives model estimation towards similar optimal parameter values as performing EM estimation directly on the full dataset.
However, compared to direct EM estimation, VHEM-H3M is more memory- and time-efficient.
First, it no longer requires storing in memory the entire data set during parameter estimation. Second, it does not need to evaluate the likelihood of all the samples at each iteration, and converges to effective estimates in shorter times. 
Note that even if a parallel implementation of EM could effectively handle the high memory requirements, a parallel-VHEM will still require fewer resources than a parallel-EM. 

In addition, for the hierarchical procedure, the estimation of the intermediate models can be easily parallelized, since they are learned independently of each other.
\ted{Finally, hierarchical estimation allows for efficient model updating when adding new data.  Assuming that the previous intermediate models have been saved, re-estimating the H3M requires learning the intermediate models of only the new data, followed by running VHEM again.  Since estimation of the intermediate models is typically as computationally intensive as the VHEM stage, reusing the previous intermediate models will lead to considerable computational savings when re-estimating the H3M.} 

In hierarchical estimation (EM on each time-series, VHEM on intermediate models), VHEM implicitly averages over all possible observations (virtual variations of each time-series) compatible with the intermediate models. We expect this to regularize estimation, which may result in models that generalize better (compared to estimating models with direct EM).
Lastly, the ``virtual'' samples (i.e., sequences), which VHEM implicitly generates for maximum-likelihood estimation, need not be of the same length as the actual input data for estimating the intermediate models. Making the virtual sequences relatively short will positively impact the run time of each VHEM iteration. 
	This may be achieved without loss of modeling accuracy,  
	as show in Section \ref{sec:robustness}.


\subsection{Related work}\label{sec:related}

\cite{jebara2007}'s approach to clustering HMMs consists of applying spectral clustering to a probability product kernel (PPK) matrix between HMMs --- we will refer to it as PPK-SC. In particular, the PPK similarity between two HMMs, $\calM\s{a}$ and $\calM\s{b}$, is defined as
\begin{equation}
\begin{split}
k(a,b) =& \int p(y_{1:\tau}|\calM\s{a})^\lambda \, p(y_{1:\tau}|\calM\s{b})^\lambda dy_{1:\tau},
\end{split}
\label{eg:ppk}
\end{equation}
where $\lambda$ is a scalar, and $\tau$ is the length of ``virtual'' sequences. The case $\lambda = \frac{1}{2}$ corresponds to the Bhattacharyya affinity.
While this approach indirectly leverages the probability distributions represented by the HMMs (i.e., the PPK affinity is computed from the probability distributions of the HMMs) and has proven successful in grouping HMMs into similar clusters~\citep{jebara2007},
it has several limitations.
First, 
the spectral clustering algorithm
cannot produce novel HMM cluster centers to represent the clusters, which is suboptimal for several applications of HMM clustering. 
For example, when implementing hierarchical clustering in the spectral embedding space (e.g., using hierarchical k-means clustering), clusters are represented by \emph{single} points in the embedding space.
This may fail to capture information on the local structure of the clusters that, when using VHEM-H3M, would be encoded by the novel HMM cluster centers.
Hence, we expect VHEM-H3M to produce better hierarchical clustering than the spectral clustering algorithm, especially at higher levels of the hierarchy. This is because, when building a new level, VHEM 
can leverage more information from the lower levels,
as encoded in the HMM cluster centers.

One simple extension of PPK-SC to obtain a HMM cluster center is to select the input HMM that the spectral clustering algorithm maps  closest to the spectral clustering center.
However, with this method, the HMM cluster centers are limited to be one of the {\em existing} input HMMs (i.e., similar to k-medoids~\citep{rousseeuw1987clustering}), instead of the HMMs that optimally condense the structure of the clusters. Therefore, we expect the \emph{novel} HMM cluster centers learned by VHEM-H3M to better represent the clusters. 
A more involved, ``hybrid'' solution is to learn the HMM cluster centers with VHEM-H3M \emph{after} obtaining clusters with PPK-SC --- using the VHEM-H3M algorithm to summarize all the HMMs within each PPK-SC cluster into a single HMM.
However, 
 we expect our VHEM-H3M algorithm to learn more accurate clustering 
models, since it jointly learns the clustering and the HMM centers, by optimizing a single objective function (i.e., the lower bound to the expected log-likelihood in \refeqn{eqn:LB-cost-summary}). 

A second drawback of the spectral clustering algorithm is that the construction and the inversion of the similarity matrix between the input HMMs is a costly operation when their number is large (e.g., see the experiment on H3M density estimation on the music data in Section \ref{sec:exp:mus}).
Therefore, we expect VHEM-H3M to be computationally more efficient than the spectral clustering algorithm since, by \emph{directly} operating on the probability distributions of the HMMs, it does not require the construction of an initial embedding or any costly matrix operation on large kernel matrices.

Finally, as  \cite{jebara2004probability} note, the exact computation of \refeqn{eg:ppk} cannot be carried out efficiently, unless  $\lambda = 1$.
For different values of $\lambda$,\footnote{The experimental results in \citep{jebara2004probability} and \citep{jebara2007} suggest to use $\lambda < 1$.}  \cite{jebara2004probability} propose to \emph{approximate} $k(a,b)$ with an alternative kernel function that can be efficiently computed; this alternative kernel function, however, is not guaranteed to be invariant to different but equivalent representations of the hidden state process \citep{jebara2004probability}.%
\footnote{The kernel in \refeqn{eg:ppk} is computed by marginalizing out the hidden state variables, i.e.,  $ \int \left(\sum_{x_{1:\tau}} p(y_{1:\tau},x_{1:\tau}|\calM\s{a})\right)^\lambda \, \left(\sum_{x_{1:\tau}}p(y_{1:\tau},x_{1:\tau}|\calM\s{b})\right)^\lambda dy_{1:\tau}$. This can be efficiently solved with the junction tree algorithm only when $\lambda = 1$.
For  $\lambda \neq 1$, \cite{jebara2004probability} propose to use an alternative kernel $\tilde k$ that applies the power operation to 
the terms of the sum rather than the entire sum, where the terms are joint probabilities
$p(y_{1:\tau},x_{1:\tau})$. 
I.e., $ \tilde k(a,b) = \int \sum_{x_{1:\tau}} \left(p(y_{1:\tau},x_{1:\tau}|\calM\s{a})\right)^\lambda \, \sum_{x_{1:\tau}}\left(p(y_{1:\tau},x_{1:\tau}|\calM\s{b})\right)^\lambda dy_{1:\tau}$.
}

Note that spectral clustering algorithms similar to \citep{jebara2007} can be applied to 
kernel (similarity) matrices that are based on other affinity scores between HMM distributions than the PPK similarity of \cite{jebara2004probability}. 
Examples 
can be found in earlier work on HMM-based clustering of time-series, such as \cite{juang1985}, \cite{lyngso1999metrics}, \cite{bahlmann2001measuring}, \cite{panuccio2002hidden}.
In particular, \cite{juang1985} propose to approximate the (symmetrised) log-likelihood between two HMM distributions by computing the log-likelihood of real samples generated by one model under the other.\footnote{For two HMM distributions, $\calM\s{a}$ and $\calM\s{b}$, \cite{juang1985} consider the affinity $L(a,b) = \frac{1}{2} \left[ \log p(Y_b|\calM\s{a}) +  p(Y_a|\calM\s{b}) \right]$, where $Y_a$ and $Y_b$ are sets of observation sequences generated from $\calM\s{a}$ and $\calM\s{b}$, respectively.}
Extensions of \cite{juang1985} have been proposed by \cite{zhong2003unified} and \cite{yin2005integrating}. 
In this work we do not pursue a comparison of the various similarity functions, but implement spectral clustering only based on PPK similarity (which \cite{jebara2007} showed to be superior).

HMMs can also be clustered by sampling a number of time-series from each of the HMMs in the base mixture, and then applying the EM algorithm for H3Ms \citep{smyth1997}, to cluster the time-series.
Despite its simplicity, this approach would suffer from high memory and time requirements, especially when dealing with a large number of input HMMs. First, all generated samples need to be stored in memory.
Second, evaluating the likelihood of the generated samples at each iteration 
is computationally intensive,
and prevents the EM algorithm from converging to effective estimates in acceptable times.\footnote{In our experiments, EM on generated samples took two orders of magnitude more time than VHEM.}
On the contrary, VHEM-H3M is more efficient in computation and memory usage, as it replaces a costly sampling step (along with the associated likelihood computations at each iteration) with an expectation. 
An additional problem of EM with sampling is that, with a simple application of the EM algorithm, time-series generated from the same input HMM can be assigned to different clusters of the output H3M.
As a consequence, the resulting clustering is not necessary \emph{consistent}, since in this case the corresponding input HMM may not be clearly assigned to any single cluster.
In our experiments, we circumvent this problem by defining appropriate constrains on the assignment variables.


The VHEM algorithm is similar in spirit to Bregman-clustering in \cite{banerjee2005clustering}. Both algorithms base clustering on KL-divergence --- the KL divergence and the expected log-likelihood  differ only for an entropy term that does not affect the clustering.
\ted{The main differences are: 1) in our setting, the expected log-likelihood (and KL divergence) is not computable in closed form, and hence VHEM uses an approximation; 2) VHEM-H3M clusters random {\em processes} (i.e., time series models), whereas \citep{banerjee2005clustering} is limited to single random variables.}

In the next two sections, we validate the points raised in this discussion through experimental evaluation using the VHEM-H3M algorithm.
In particular, we consider clustering experiments in Section  \ref{sec:exp:cluster}, 
and H3M density estimation for automatic annotation and retrieval in Section \ref{sec:exP:ann}. Each application exploits some of the benefits of VHEM.
First, we show that VHEM-H3M is more accurate in clustering than PPK-SC, in particular at higher levels 
%
%
%
of a hierarchical clustering (Section \ref{sec:exP:hmc}), and in an experiment with synthetic data (Section \ref{sec:exP:synth}). 
%
%
Similarly, the annotation and retrieval results in Section \ref{sec:exP:ann} favor VHEM-H3M over PPK-SC and over standard EM, suggesting that VHEM-H3M is more robust and effective for H3M density estimation.
Finally, in all the experiments, the running time of VHEM-H3M compares favorably with the other HMM clustering algorithms; PPK-SC suffers long delays when the number of input HMMs is large 
and the standard EM algorithm is considerably slower. This demonstrates that VHEM-H3M is most efficient for clustering HMMs.

\section{Clustering Experiments}\label{sec:exp:cluster}

In this section, we present an empirical study of the VHEM-H3M algorithm for clustering and hierarchical clustering of HMMs.
Clustering HMMs consists in partitioning $K_1$ input HMMs into $K_2<K_1$ groups of similar HMMs. 
Hierarchical clustering involves organizing the input HMMs in a multi-level hierarchy with $h$ levels, by applying clustering in a recursive manner.
Each level  $\ell$ of the hierarchy has $K_\ell$ groups (with $K_1>K_2>\dots>K_{h-1}>K_h$), and
the first level consists of the $K_1$ input HMMs. 

We begin with an experiment on hierarchical clustering, where each of the input HMMs to be clustered is estimated on a sequence of motion capture data (Section \ref{sec:exP:hmc}).
Then, we present a simulation study on clustering synthetic HMMs (Section \ref{sec:exP:synth}).
First, we provide an overview of the different algorithms used in this study.

\subsection{Clustering methods}


In the clustering experiments, we will compare 
our VHEM-H3M algorithm with several other clustering algorithms.  The various algorithms are summarized here.


\begin{itemize}
\item {\bf VHEM-H3M}: 
We cluster $K_1$ input HMMs into $K_2$ clusters by using the VHEM-H3M algorithm (on the input HMMs) to learn a H3M with $K_2$ components (as explained in Section \ref{subsec:form}).
To build a multi-level hierarchy of HMMs with $h$ levels, we start from the first level of $K_1$ input HMMs, and  recursively use the VHEM-H3M algorithm $h-1$ times.
Each new level $\ell$ is formed by clustering the $K_{\ell-1}$ HMMs at the previous level into $K_{\ell} < K_{\ell-1}$ groups with the VHEM-H3M algorithm, and using the learned HMMs as cluster centers at the new level. In our experiments, we set the number of virtual samples to $N = {10}^{4} K\s{\ell-1}$, a large value that favors ``hard'' clustering (where each HMM is univocally assigned to a single cluster), 
and the length of the virtual sequences to $\tau = 10$.

\item {\bf PPK-SC}: \ted{\cite{jebara2007} cluster HMMs by calculating a PPK similarity matrix between all HMMs, and then applying spectral clustering.}
\ted{The work in \cite{jebara2007} only considered HMMs with single Gaussian emissions, which did not always give satisfactory results in our experiments.  Hence, we extended \citep{jebara2007} by allowing GMM emissions,
and derived the PPK similarity for this more general case using \citep{jebara2004probability}.}
From preliminary experiments, we found the best performance for PPK  with
$\lambda = \frac{1}{2}$ (i.e., Bhattacharyya affinity), and when integrating over sequences of length $\tau = 10$.
%
Finally, we also extend \cite{jebara2007} to construct multi-level hierarchies, by using hierarchical k-means in the spectral clustering embedding.

\item {\bf SHEM-H3M}: \ted{This is a version of HEM-H3M that maximizes the likelihood of {\em actual} samples generated from the input HMMs, as in \refeqn{eqn:LL}, rather than the expectation of virtual samples, as in \refeqn{eqn:cost_function}.}
In particular, from each input HMM $\calM\s{b}_i$ we sample a set $Y_i$ of $N_i = \pi\s{b}_i N$ observation sequences (for a large value of $N$).
We then estimate the reduced H3M from the $N$ samples $Y = \{Y_i\}_{i=1}^{K\s{b}}$, with the EM-H3M algorithm of \cite{smyth1997}, 
which was modified to use a single assignment variable for each sample set $Y_i$, to obtain a consistent clustering.

\end{itemize}

In many real-life applications, the goal is to cluster a collection of time series, i.e., observed sequences.
Although the input data is not a collection of HMMs in that case, it can still be clustered with the VHEM-H3M algorithm by first modeling each sequence as an HMM, and then using the HMMs as input for the VHEM-H3M algorithm.
With time-series data as input, it is also possible to use clustering approaches that do not model each sequence as a HMM.
Hence, in one of the hierarchical motion clustering experiments, we also compare to the following two algorithms, one that clusters time-series data directly \citep{smyth1997}, and a second one that clusters the time series after modeling each sequence with a dynamic texture (DT) model \citep{Chan2010cvpr}.

\begin{itemize}

\item {\bf EM-H3M}: The EM algorithm for H3Ms \citep{smyth1997}  is applied directly on a collection of time series to learn the clustering and HMM cluster centers, thus bypassing the intermediate HMM modeling stage. 
To obtain a hierarchical clustering (with $h\geq 3$ levels), we proceed in a bottom up fashion and build each new level 
by simply re-clustering the given time series in a smaller number of clusters using \citep{smyth1997}. We extend the algorithm to use a single assignment variable for each set of sequences $Y_i$
that  are within the same cluster in the immediately lower level of the hierarchy. 
This modification preserves the hierarchical clustering property that sequences in a cluster will remain together at the higher levels.

\item {\bf HEM-DTM}: \ted{Rather than use HMMs, we consider 
a clustering model based on
linear dynamical systems, i.e., dynamic textures (DTs) \citep{Doretto2003}. 
Hierarchical clustering is performed using the
hierarchical EM algorithm for DT mixtures (HEM-DTM) \citep{Chan2010cvpr}, in an analogous way to VHEM-H3M.
The main difference is that, with HEM-DTM, time-series are modeled as 
DTs, which have a {\em continuous} state space (a Gauss-Markov model) and {\em unimodal} observation model, whereas VHEM-H3M uses a {\em discrete} state space and {\em multimodal} observations (GMMs).} 


\end{itemize}

We will use several metrics to quantitatively compare the results of different clustering algorithms. 
First, we will calculate the {\em Rand-index} \citep{hubert1985comparing}, which measures the
\ted{correctness} of a proposed clustering against a given ground truth clustering. Intuitively, this index measures how consistent cluster assignments are with the ground truth (i.e., whether pairs of items are correctly or incorrectly assigned to the same cluster, or different clusters).
Second, we will consider the {\em log-likelihood}, as used by \cite{smyth1997} to evaluate a clustering. This measures how well the clustering fits the input data.
When time series are given as input data, we compute the log-likelihood of a clustering as the sum of the log-likelihoods of each input sequence
under the HMM cluster center to which it has been assigned.
When the input data consists of HMMs, we will evaluate the log-likelihood of a clustering by using the expected log-likelihood of observations generated from an input HMM under the HMM cluster center to which it is assigned.
For PPK-SC, the cluster center is estimated by running the VHEM-H3M algorithm (with $K\s{r}=1$) on the HMMs assigned to the cluster.\footnote{Alternatively, we could use as cluster center the HMM mapped the closest to the spectral embedding cluster center, but this always resulted in lower log-likelihood.}
Note that the log-likelihood will be particularly appropriate to compare VHEM-H3M, SHEM-H3M, EM-H3M and HEM-DTM, since they explicitly optimize for it. However, it may be unfair for PPK-SC, since this method optimizes the PPK similarity and not the log-likelihood.
As a consequence, we also measure
the {\em PPK cluster-compactness}, which is more directly related to what PPK-SC optimizes for. The PPK cluster-compactness is the sum (over all clusters) of the average intra-cluster PPK pair-wise similarity. This performance metric favors methods that produce clusters with high intra-cluster similarity.

\subsection{Hierarchical motion clustering}\label{sec:exP:hmc}

\begin{figure}[tb]

\centering
  \renewcommand{\arraystretch}{1.3}
  \begin{tabular}{c@{\hspace{.1in}}c@{\hspace{.1in}}c@{\hspace{.1in}}c}
  \multicolumn{4}{c}{\scriptsize (a) ``Walk'' sequence.}\\
      \includegraphics[width=1.2in,height =.8in]{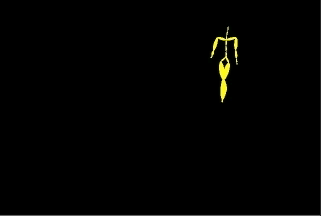} &   \includegraphics[width=1.2in,height =.8in]{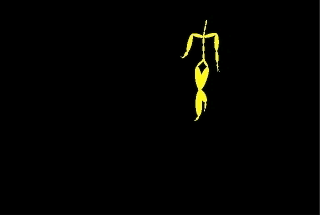} &\includegraphics[width=1.2in,height =.8in]{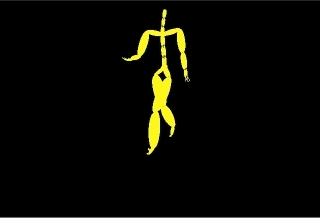} &\includegraphics[width=1.2in,height =.8in]{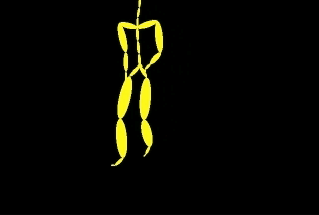}\\
            \multicolumn{4}{c}{\scriptsize (b) ``Run'' sequence.}\\
            \includegraphics[width=1.2in,height =.8in]{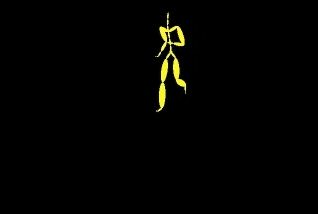} &   \includegraphics[width=1.2in,height =.8in]{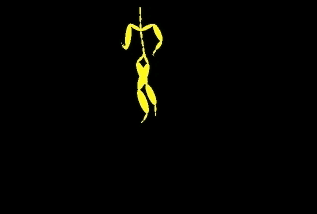} &\includegraphics[width=1.2in,height =.8in]{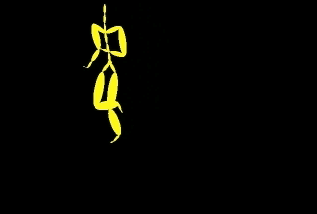} &\includegraphics[width=1.2in,height =.8in]{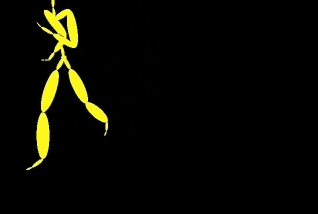}\\
  \end{tabular}
\caption{Examples of motion capture sequences from the MoCap dataset, shown with stick figures.
}
\label{fig:stick}
\end{figure}

\begin{figure}[h]
\begin{center}
\vspace{-.09in}
\includegraphics[width=0.85\textwidth]{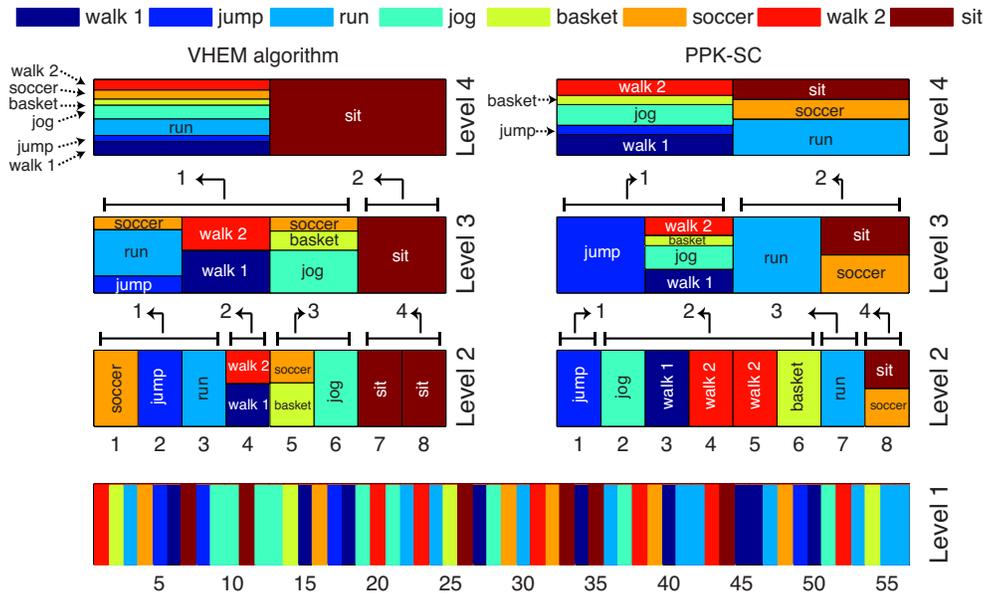}
\end{center}
\vspace{-.15in}
\caption{An example of hierarchical clustering of the MoCap dataset, with VHEM-H3M and PPK-SC. 
Different colors represent different motion classes. 
Vertical bars represent clusters, with the colors indicating the proportions of the motion classes in a cluster, and the numbers on the x-axes representing the clusters' indexes.
At Level 1 there are $56$ clusters, one for each motion sequence. At Levels $2$, $3$ and $4$ there are $8$, $4$ and $2$ HMM clusters, respectively.
For VHEM almost all clusters at Level 2 are populated by examples from a single motion class. The error of VHEM in clustering a portion of ``soccer''  with  ``basket'' is probably because both actions involve a sequence of movement, shot, and pause. Moving up the hierarchy, the VHEM algorithm clusters similar motions classes together, and at Level 4 creates a dichotomy between ``sit'' and the other (more dynamic) motion classes.
 PPK-SC also clusters motion sequences well at Level 2, but incorrectly aggregates ``sit'' and ``soccer'', which have quite different dynamics.  At Level 4,  the clustering obtained by PPK-SC is harder to interpret than that by VHEM.
}
\label{fig:hier}
\end{figure} 

In this experiment we test the VHEM algorithm on hierarchical motion clustering
from motion capture data, i.e.,  time series representing human locomotions and actions.
 %
 %
To hierarchically cluster a collection of time series, we first model each time series with an HMM and then cluster the HMMs hierarchically.
Since each HMM summarizes the appearance and dynamics of the particular motion sequence it represents, the structure encoded in the hierarchy of HMMs directly applies to the original motion sequences.
\cite{jebara2007} uses a similar approach to cluster motion sequences, applying PPK-SC to cluster HMMs. However, they did not extend their study to hierarchies with multiple levels.

\subsubsection{Datasets and setup}

\ted{We experiment on two motion capture datasets,
the MoCap dataset (\url{http://mocap.cs.cmu.edu/}) and the Vicon Physical Action dataset \citep{theodoridis2007action,Frank+Asuncion:2010}.}
For the MoCap dataset, we use $56$ motion examples spanning $8$ different classes (``jump'', ``run'', ``jog'', ``walk 1'',  ``walk 2'', 
``basket'', ``soccer'', and ``sit''). Each example is a sequence of $123$-dimensional vectors representing the $(x,y,z)$-coordinates of $41$ body markers tracked spatially through time. Figure \ref{fig:stick} illustrates some typical examples.
We built a hierarchy of $h = 4$ levels.
The first level (Level 1) was formed by the $K_1 = 56$ HMMs learned from each individual motion example (with $S = 4$ hidden states, and $M = 2$ components for each GMM emission).
The next three levels contain $K_2 = 8$, $K_3 = 4$ and $K_4 = 2$ HMMs. 
We perform the hierarchical clustering with VHEM-H3M, PPK-SC, EM-H3M, SHEM-H3M ($N \in \{560, 2800\}$ and $\tau = 10$), and HEM-DTM (state dimension of $7$).
The experiments were repeated $10$ times for each clustering method, 
using different random initializations of the algorithms.

The Vicon Physical Action dataset is a collection of $200$ motion sequences. Each sequence consists of a time series of $27$-dimensional vectors representing the $(x,y,z)$-coordinates of $9$ body markers captured using the Vicon 3D tracker. The dataset includes $10$ normal and $10$ aggressive activities, performed by each of $10$ human subjects a single time. We build a hierarchy of $h = 5$ levels, 
 starting with $K_1 = 200$ HMMs (with $S = 4$ hidden states and $M=2$ components for each GMM emission) at the first level (i.e., one for each motion sequence), and using $K_2 = 20$, $K_3 = 8$, $K_4 = 4$, and $K_5 = 2$ for the next four levels.
The experiment was repeated $5$ times 
with VHEM-H3M and PPK-SC, using different random initializations of the algorithms.

In similar experiments where we varied the number of levels $h$ of the hierarchy and the number of clusters at each level, we noted similar relative performances of the various clustering algorithms, on both datasets.

\subsubsection{Results on the MoCap dataset}

An example of hierarchical clustering of the MoCap dataset with VHEM-H3M is illustrated in Figure \ref{fig:hier} (left).
In the first level, each vertical bar represents a motion sequence, with different colors indicating different ground-truth classes.
In the second level, the $K_2 = 8$ HMM clusters are shown with vertical bars, with the colors indicating the proportions of the motion classes in the cluster.
Almost all clusters are populated by examples from a single motion class (e.g., ``run'', ``jog'', ``jump''), which demonstrates that VHEM can group similar motions together. 
We note an error of VHEM in clustering a portion of the ``soccer'' examples with  ``basket''. This is probably caused by the similar dynamics of these actions, which both consist of a sequence of movement, shot, and pause.
Moving up the hierarchy, the VHEM algorithm clusters similar motion classes together (as indicated by the arrows), for example ``walk 1'' and ``walk 2'' are clustered together at Level 2, and at the highest level (Level 4) it creates a dichotomy between ``sit'' and the rest of the motion classes. This is a desirable behavior  as the kinetics of the ``sit'' sequences (which in the MoCap dataset correspond to starting in a standing position, sitting on a stool, and returning to a standing position)
are considerably different from the rest.
On the right of Figure \ref{fig:hier}, the same experiment is repeated with PPK-SC.
PPK-SC clusters motion sequences properly, but incorrectly aggregates ``sit'' and ``soccer'' at Level 2, even though they have quite different dynamics.
Furthermore, the highest level (Level 4) of the hierarchical clustering produced by PPK-SC is harder to interpret than that of VHEM.

	\begin{table}[t]
	\caption{Hierarchical clustering of the MoCap dataset using VHEM-H3M, PPK-SC, SHEM-H3M, EM-H3M and HEM-DTM.
 The number in brackets after SHEM-H3M represents the number of real samples used.
	We computed Rand-index, data log-likelihood and cluster compactness at each level of the hierarchy,
	 and registered the time (in seconds) to learn the hierarchical structure.
Differences in Rand-index at Levels 2, 3, and 4 are statistically significant based on a paired t-test with confidence $95\%$. 
	 }
		\label{tab:mocap}
	\begin{center}
	\resizebox{\columnwidth}{!}{
	\begin{tabular}{l|ccc|ccc|ccc|c|}
 & \multicolumn{3}{c|}{Rand-index}  & \multicolumn{3}{c|}{log-likelihood ($\times 10^6$)} &\multicolumn{3}{c|}{PPK cluster-compactness} & time (s) \\
\multicolumn{1}{r|}{Level} & 2 & 3 & 4 & 2 & 3 & 4  & 2 & 3 & 4 &    \\
	\hline 
VHEM-H3M 	&     0.937  &  \bf{0.811} &   \bf{0.518}  & \bf{-5.361}  	&  \bf{-5.682} 	&  \bf{-5.866}   		& 0.0075	& \bf{0.0068} 	& \bf{0.0061}  & \bf{30.97}\\
PPK-SC 		&     \bf{0.956}   & 0.740  &  0.393  & -5.399 	&  -5.845 	&  -6.068 			& 0.0082	& 0.0021 & 0.0008 &  37.69\\
SHEM-H3M (560)& 0.714  &  0.359  &  0.234  & -13.632  &  -69.746 &  -275.650 & 0.0062  &  0.0034  &  0.0031 &843.89 \\
SHEM-H3M (2800)& 0.782  &  0.685  &  0.480 &  -14.645 & -30.086 & -52.227 & 0.0050  &  0.0036  &  0.0030  &  3849.72 \\
 EM-H3M 	&     0.831  &  0.430  &  0.340  & -5.713  	&  -202.55 	&  -168.90   		& \bf{0.0099}  &  0.0060  &  0.0056  & 667.97 \\
{HEM-DTM}	&    {0.897}	  &{0.661} &   {0.412} & {-7.125}  & {-8.163}    &  {-8.532} & - & - & - & 121.32
	\end{tabular}
	}
	\end{center}
	\vspace{-.2in}
	\end{table}

Table \ref{tab:mocap} presents a quantitative comparison between PPK-SC and VHEM-H3M at each level of the hierarchy.
While VHEM-H3M has lower Rand-index than PPK-SC at Level 2 ($0.937$ vs. $0.956$), VHEM-H3M 
has higher Rand-index at Level 3 ($0.811$ vs. $0.740$) and Level 4 ($0.518 $ vs. $0.393 $).
In terms of PPK cluster-compactness, we observe similar results.
In particular, VHEM-H3M has higher PPK cluster-compactness than PPK-SC at Level 3 and 4. Overall, keeping in mind that PPK-SC is explicitly driven by PPK-similarity, while the VHEM-H3M algorithm is not, these results can be considered as strongly in favor of VHEM-H3M (over PPK-SC).
In addition, the data log-likelihood for VHEM-H3M is higher than that for PPK-SC at each level of the hierarchy. This suggests that the novel HMM cluster centers learned by VHEM-H3M fit the motion capture data better 
than the spectral cluster centers. 
 This conclusion is further supported by the results of the density estimation experiments
 in Sections \ref{sec:exp:mus} and \ref{sec:exp:dig}.
Note that the higher up in the hierarchy, the more clearly this effect is manifested.

Comparing to other methods (also in Table \ref{tab:mocap}),
EM-H3M generally has lower Rand-index than VHEM-H3M and PPK-SC (consistent with the results in \citep{jebara2007}).
While EM-H3M directly clusters the original motion sequences, both VHEM-H3M and PPK-SC implicitly integrate over all possible virtual variations of the original motion sequences (according to the intermediate HMM models), which results in more robust clustering procedures.
In addition, EM-H3M has considerably longer running times than VHEM-H3M and PPK-SC (i.e., roughly $20$ times longer) since it needs to evaluate the likelihood of all training sequences at each iteration, at all levels.

The results in Table \ref{tab:mocap} favor VHEM-H3M over SHEM-H3M, and
empirically validate the variational approximation that VHEM uses for learning.
For example, when using $N = 2800$ samples, running SHEM-H3M takes over two orders of magnitude more time than VHEM-H3M, but still does not achieve performance competitive with VHEM-H3M.
With an efficient closed-form expression for averaging over all possible virtual samples, VHEM approximates the sufficient statistics of a virtually unlimited number of observation sequences, without the need of using real samples. This has an additional regularization effect that improves the robustness of the learned HMM cluster centers.
In contrast, SHEM-H3M uses real samples, and requires a large number of them to learn accurate models, which results in significantly longer running times.

Finally, in Table \ref{tab:mocap}, we also report hierarchical clustering performance for HEM-DTM.  
VHEM-H3M consistently outperforms 
HEM-DTM, both in terms of Rand-index and data log-likelihood.\footnote{We did not report PPK cluster-compactness for HEM-DTM, since it would not be directly comparable with the same metric based on HMMs.} 
Since both VHEM-H3M and HEM-DTM are based on the hierarchical EM algorithm for learning the clustering, 
this indicates that HMM-based clustering models are more appropriate than DT-based models for the human MoCap data.
Note that, while PPK-SC is also HMM-based, it has a lower  Rand-index than HEM-DTM at Level 4.
This further suggests that PPK-SC does not optimally 
cluster the HMMs. 

	\begin{table}[t]
	\caption{Hierarchical clustering of the Vicon Physical Action dataset using VHEM-H3M and PPK-SC. 
	Performance is measured in terms of Rand-index, data log-likelihood and PPK cluster-compactness at each level. 
Differences in Rand-index at Levels 2, 4 and 5 are statistically significant based on a paired t-test with confidence $95\%$. The test failed at Level 3. 
	}
		\label{tab:viconCluster}
	\begin{center}
	\resizebox{\columnwidth}{!}{
	\begin{tabular}{l|cccc|cccc|cccc|}
 & \multicolumn{4}{c|}{Rand-index}  & \multicolumn{4}{c|}{log-likelihood ($\times 10^6$)} &\multicolumn{4}{c|}{PPK cluster-compactness} \\
\multicolumn{1}{r|}{Level} & 2 & 3 & 4 & 5 & 2 & 3 & 4 & 5 & 2 & 3 & 4 & 5   \\
	\hline 
VHEM-H3M 	&     0.909  &  0.805  &  0.610 &   0.244 & -1.494   &  -3.820    &  -5.087   &  -6.172     & 0.224 &    0.059 &   0.020 &   0.005  \\
PPK-SC 		&     0.900   & 0.807  &  0.452  &  0.092 &  -3.857 &  -5.594  & -6.163 &   -6.643 	& 0.324&   0.081 &   0.026 &   0.008 \\
\end{tabular}
	}
	\end{center}
	\vspace{-.2in}
	\end{table}

\subsubsection{Results on the Vicon Physical Action dataset}
Table \ref{tab:viconCluster} presents results using VHEM-H3M and PPK-SC to cluster the Vicon Physical Action dataset.
While the two algorithms performs similarly 
in terms of Rand-index at lower levels of the hierarchy (i.e., Level 2 and Level 3),
at higher levels (i.e., Level 4 and Level 5) VHEM-H3M outperforms PPK-SC. 
In addition, VHEM-H3M registers higher data log-likelihood than PPK-SC at each level of the hierarchy. 
This, again, suggests that by learning new cluster centers, the VHEM-H3M algorithm retains more information on the clusters' structure than PPK-SC.
Finally, compared to VHEM-H3M, PPK-SC produces clusters that are more compact in terms of PPK similarity.
However, this does not necessarily imply a better agreement with the ground truth clustering, as evinced by 
the Rand-index metrics.



\subsection{Clustering synthetic data}\label{sec:exP:synth}

In this experiment, we compare VHEM-H3M and PPK-SC on clustering a synthetic dataset of HMMs.

\subsubsection{Dataset and setup}

\ted{
The synthetic dataset of HMMs is generated as follows.
Given a set of $C$ HMMs $\{\calM\s{c}\}_{c=1}^C$,
for each HMM we synthesize a set of $K$  ``noisy'' versions of the original HMM.}
Each ``noisy'' HMM $\tilde\calM\s{c}_k $ ($k = 1,\dots K$) is synthesized by generating a random sequence $y_{1:T}$ of length $T$ from $\calM\s{c}$, corrupting it with Gaussian noise $\sim {\cal N}(0,\sigma_n^2 \Id_d)$, and estimating the parameters of $\tilde\calM\s{c}_k $ 
on the corrupted version of $y_{1:T}$. 
%
Note that this procedure adds 
noise in the \emph{observation} space. 
%
%
%
The number of noisy versions (of each given HMM), $K$,
and the noise variance, $\sigma_n^2$, will be varied during the experiments.

The collection of original HMMs was created as follows.
Their number was always set to $C= 4$, the number of hidden states of the HMMs to $S = 3$, the emission distributions to be single, one-dimensional Gaussians (i.e., GMMs with $M = 1$ component), and the length of the sequences to $T = 100$.
For all original HMMs $\calM\s{c}$, the initial state probability and state transition matrix were fixed as
\begin{align}
 &\pi\s{c} = \left[ \begin{array}{c}  1/3 \\  1/3 \\ 1/3 \\ \end{array} \right]  , \quad 
 A\s{c} = \left[ \begin{array}{ccc}  0.8 & 0.1 & 0.1 \\   0.2 & 0.8 &  0 \\ 0 & 0.2 & 0.8  \\ \end{array} \right], \forall c.
 \end{align}
We consider two settings of the emission distributions.
In the first setting, experiment (a), the HMMs $\calM\s{c}$ only differ in the means of the emission distributions,
\begin{equation}
\begin{split}
 &\left\{ \begin{array}{c} \mu\s{1}_1 = 1        \\ \mu\s{1}_2 = 2         \\ \mu\s{1}_3 = 3        \\ \end{array}\right.,
 \left\{ \begin{array}{c} \mu\s{2}_1 = 3        \\ \mu\s{2}_2 = 2         \\ \mu\s{2}_3 = 1        \\ \end{array}\right.,
 \left\{ \begin{array}{c} \mu\s{3}_1 = 1        \\ \mu\s{3}_2 = 2         \\ \mu\s{3}_3 = 2        \\ \end{array}\right.,
  \left\{ \begin{array}{c} \mu\s{4}_1 = 1        \\ \mu\s{4}_2 = 3         \\ \mu\s{4}_3 = 3        \\ \end{array}\right.,
   \quad {\sigma\s{c}_\rho}^2 = 0.5, \quad\forall \rho,c .  
\end{split}
\label{eqn:HMM1}
\end{equation}
In the second setting, experiment (b), the HMMs differ in the variances of the emission distributions,
\begin{equation}
\begin{split}
 &\left\{ \begin{array}{c} \mu\s{c}_1 = 1        \\ \mu\s{c}_2 = 2         \\ \mu\s{c}_3 = 3        \\ \end{array}\right., \forall c,
   \quad {\sigma\s{1}_\rho}^2 = 0.5,   \quad {\sigma\s{2}_\rho}^2  = 0.1,   \quad {\sigma\s{3}_\rho}^2  = 1,   \quad {\sigma\s{4}_\rho}^2 = 0.05, \quad\forall \rho.
\end{split}
\label{eqn:HMM1}
\end{equation}

The VHEM-H3M and PPK-SC algorithms are used to cluster the synthesized HMMs, $\{\{ \tilde\calM\s{c}_k \}_{k=1}^{K}\}_{c=1}^{C}$, into $C$ groups,
and the quality of the resulting clusterings is measured with the Rand-index, 
PPK cluster-compactness, 
and the expected log-likelihood of the discovered cluster centers with respect to the original HMMs.
The expected log-likelihood was computed using the lower bound, as in 
\refeqn{eqn:LB-HMM}, with each of the original HMMs assigned to the most likely HMM cluster center. 
The results are averages over $10$ trials.



\subsubsection{Results}

\begin{figure}[tb]
\centering
  \renewcommand{\arraystretch}{1.3}
  \begin{tabular}{c@{\hspace{.1in}}c@{\hspace{.1in}}c}
  \multicolumn{3}{c}{\scriptsize Experiment (a): different  emission means.}\\
      \includegraphics[width = .3\columnwidth]{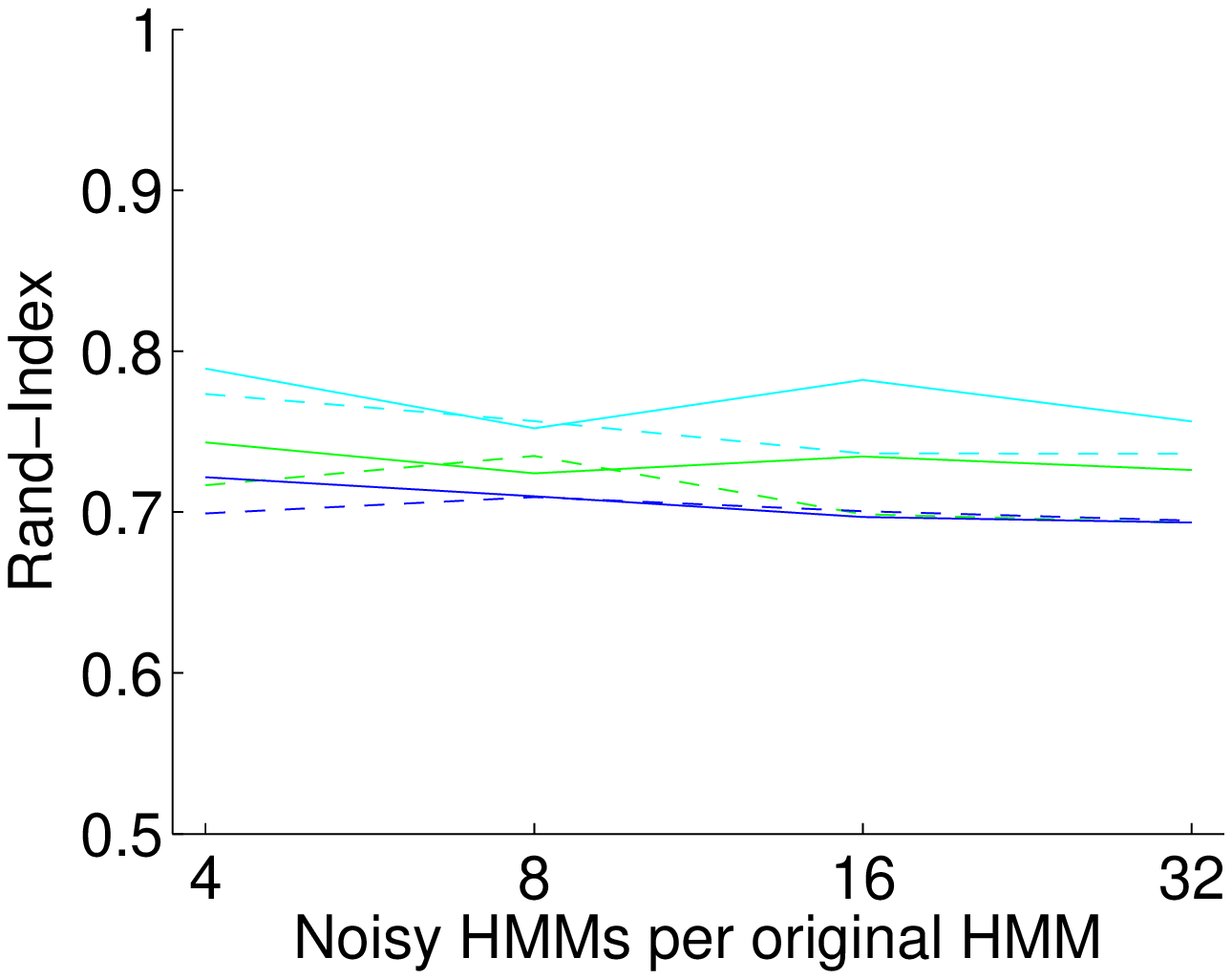} &   \includegraphics[width = .3\columnwidth]{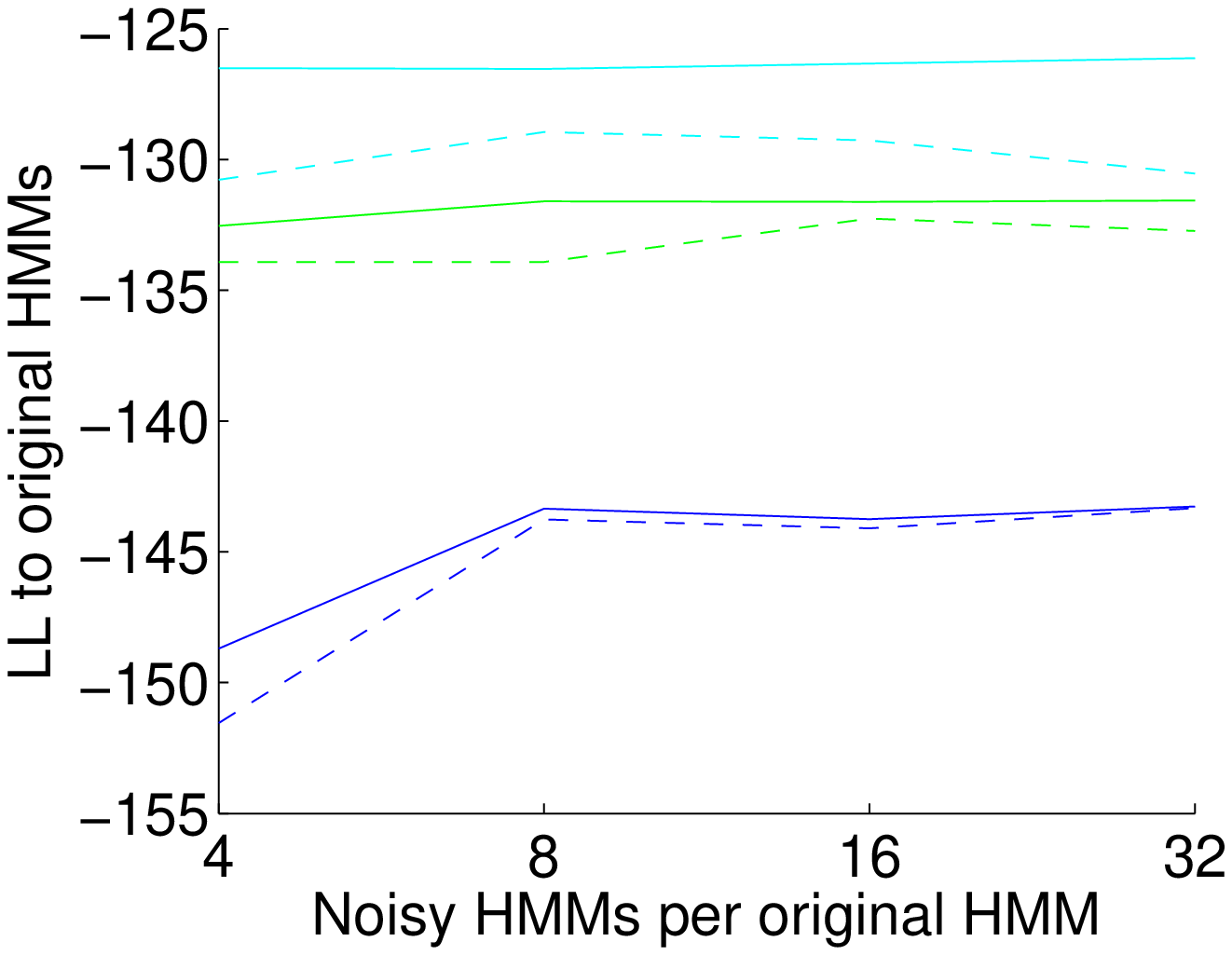}  &\includegraphics[width = .3\columnwidth]{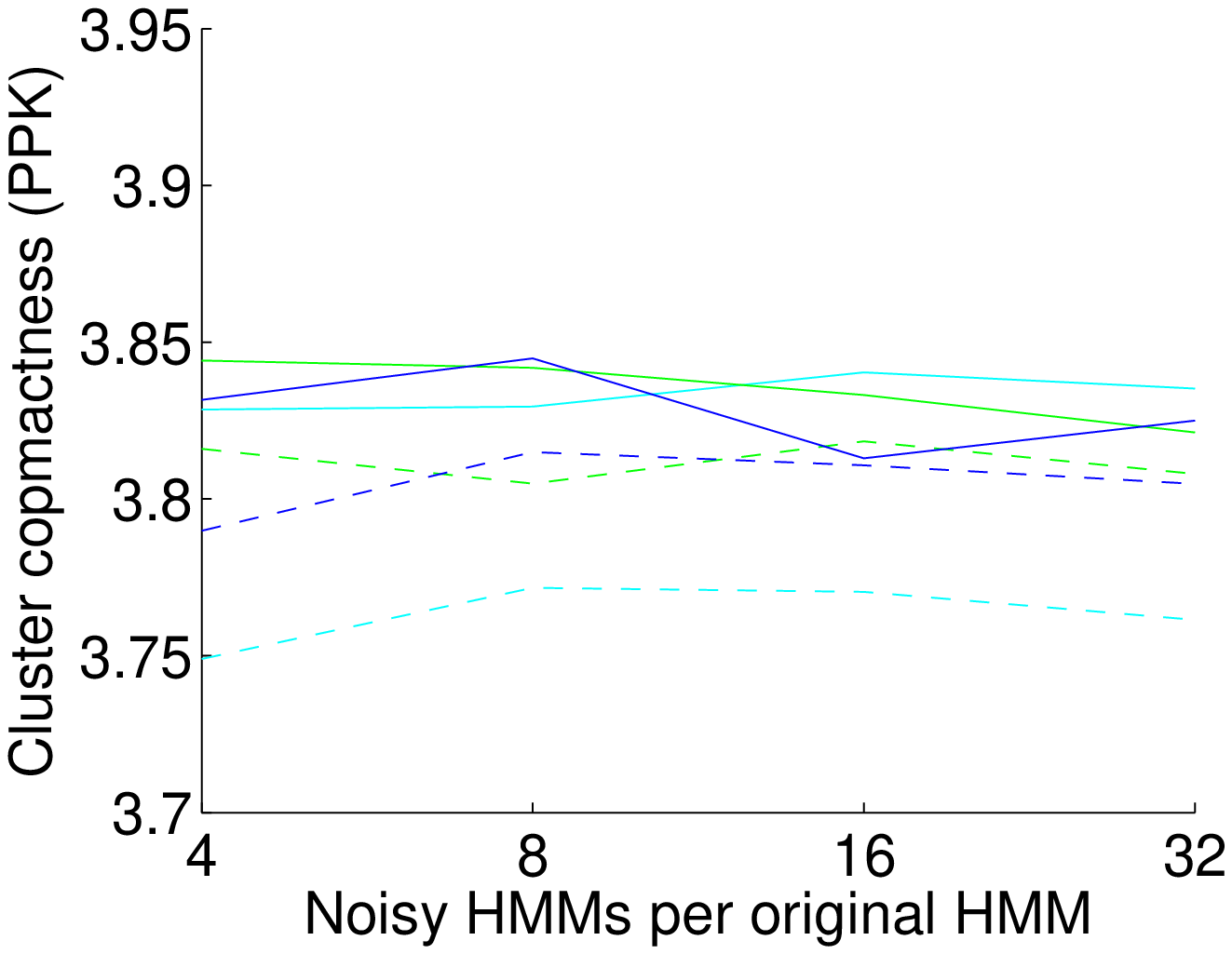} \\
            \multicolumn{3}{c}{\scriptsize  Experiment (b): different  emission variances.}\\
      \includegraphics[width = .3\columnwidth]{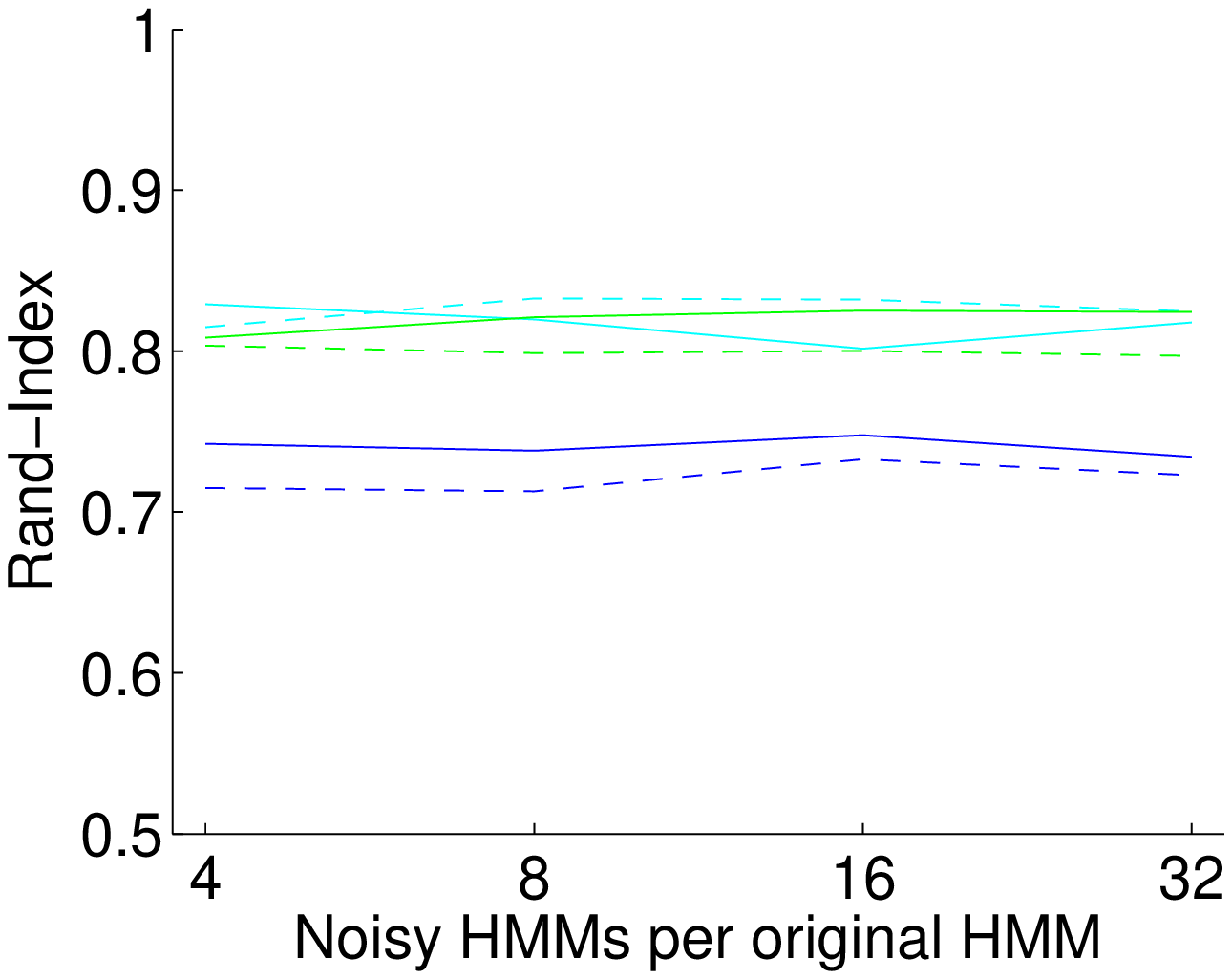} &   \includegraphics[width = .3\columnwidth]{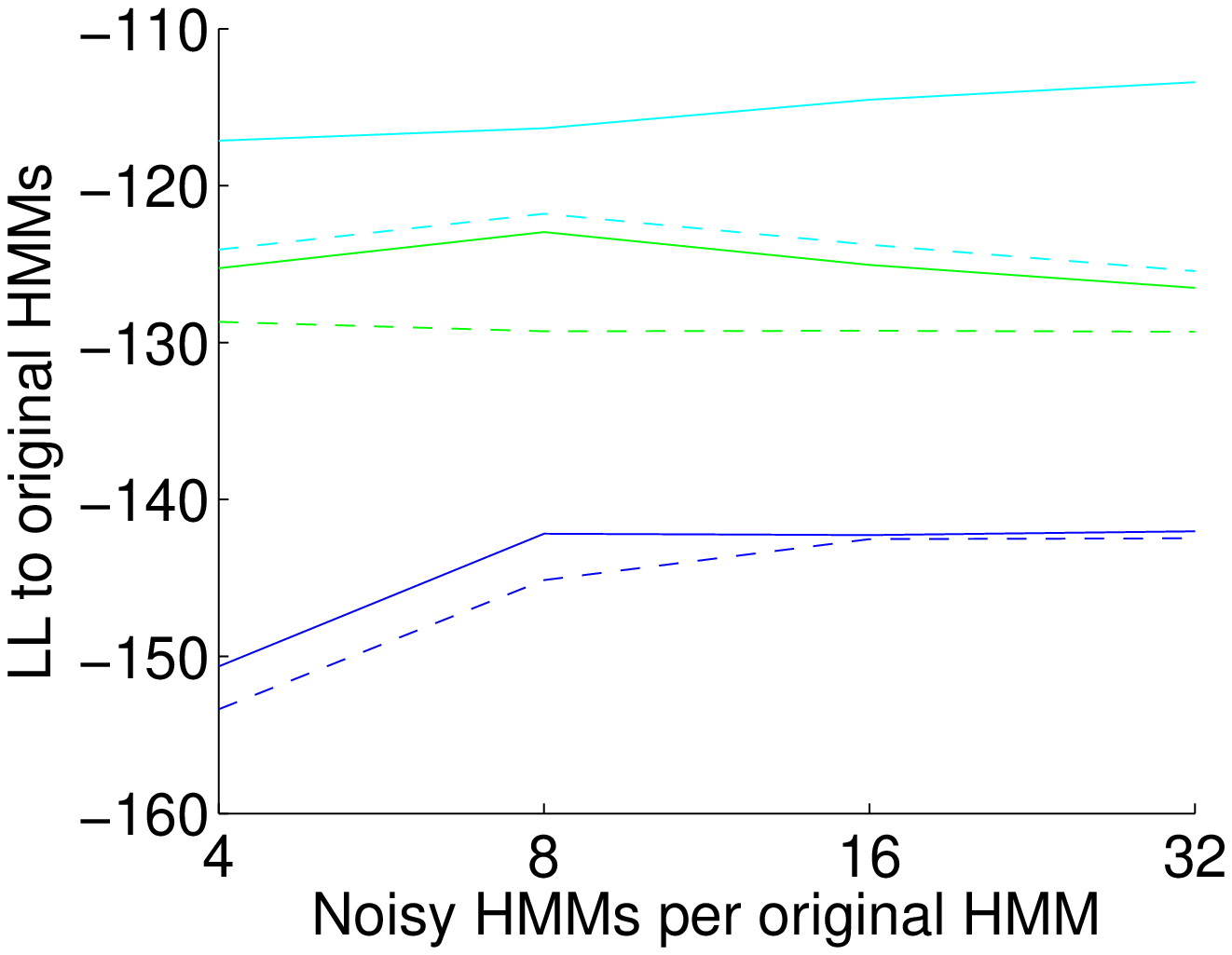}  &\includegraphics[width = .3\columnwidth]{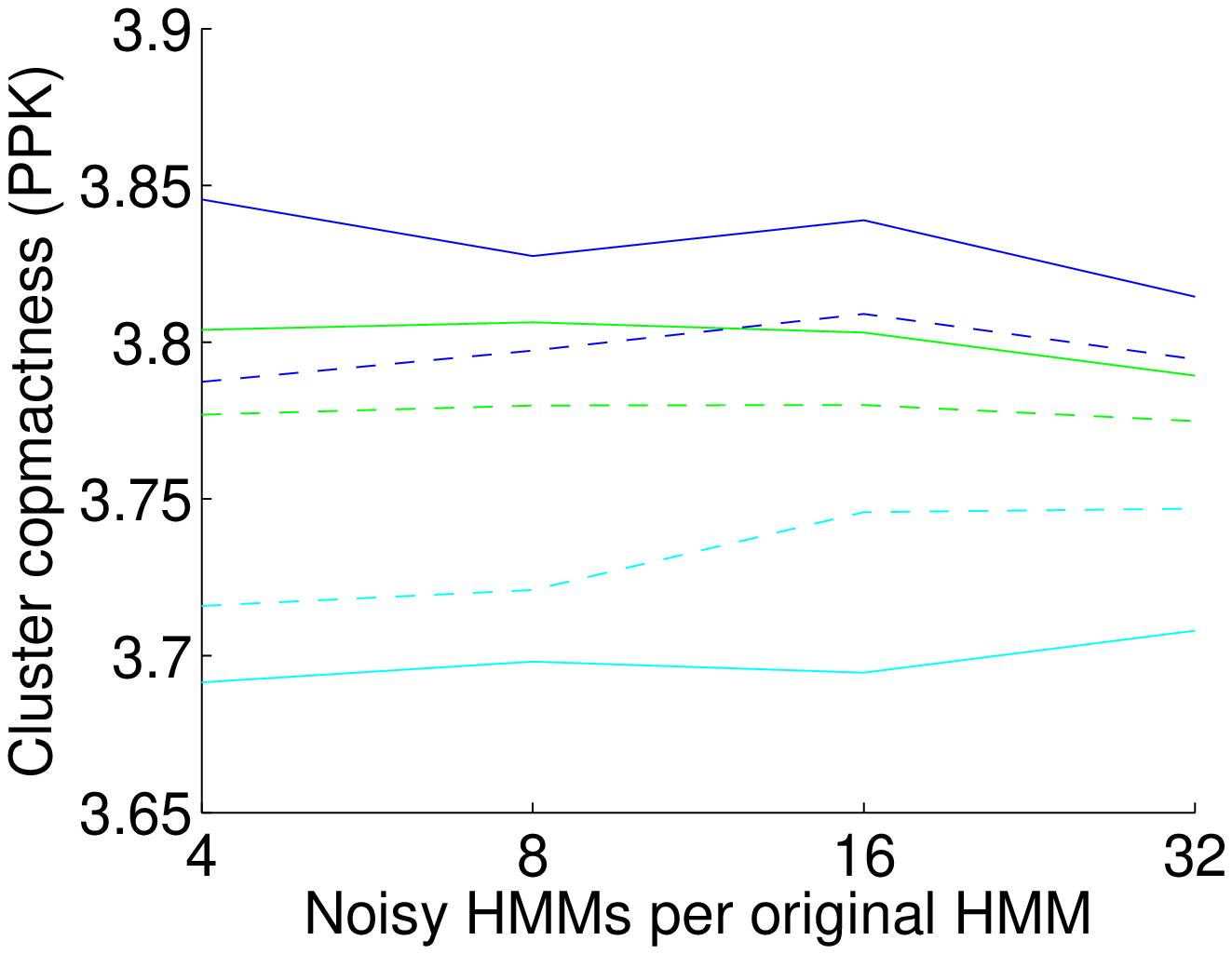} \vspace{.1in}
  \end{tabular}
  \includegraphics[width = 1\columnwidth]{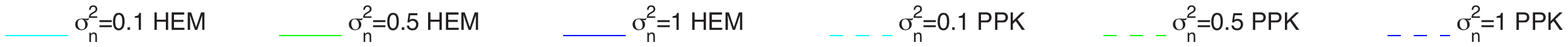}
\caption{Results on clustering synthetic data with VHEM-H3M and PPK-SC. Performance is measured in terms of Rand-index, expected log-likelihood and PPK cluster-compactness. }
\label{fig:synth}
\end{figure}

Figure \ref{fig:synth} reports the performance metrics when varying the number $K\in\{2,4,8,16,32\}$ of noisy versions of each of the original HMMs, and the noise variance $\sigma_n^2\in\{0.1,0.5,1\}$, for the two experimental settings.
%
%
%
%
For the majority of settings of $K$ and $\sigma_n^2$,  the clustering produced by VHEM-H3M is superior to the one produced by PPK-SC, for each of the considered metrics (i.e., in the plots, solid lines are usually above dashed lines of the same color). The only exception is in experiment (b) where, for low noise variance (i.e., $\sigma_n^2 = 0.1$) PPK-SC is the best in terms of Rand-index and cluster compactness.
It is interesting to note that the gap in performance between VHEM-H3M and PPK-SC is generally larger at low values of $K$. We believe this is because, when only a limited number of input HMMs is available, PPK-SC produces an embedding of lower quality. 
This does not affect VHEM-H3M, since it clusters in HMM distribution space and does not use an embedding. 

These results suggest that, by clustering HMMs {\em directly} in distribution space, VHEM-H3M is generally more robust than PPK-SC, the performance of which instead 
depends on the quality of the underlying embedding. 

\section{Density estimation experiments}
\label{sec:exP:ann}

\ted{In this section, we present an empirical study of VHEM-H3M for density estimation, in automatic annotation and retrieval of music (Section \ref{sec:exp:mus}) and hand-written digits (Section \ref{sec:exp:dig}).} 


\subsection{Music annotation and retrieval}\label{sec:exp:mus}

In this experiment, we evaluate VHEM-H3M for estimating annotation models in content-based music auto-tagging.
As a generative time-series model, H3Ms allow to account for timbre (i.e., through the GMM emission process) as well as longer term temporal dynamics (i.e., through the HMM hidden state process), when modeling musical signals. Therefore, in music annotation and retrieval applications, H3Ms are expected to prove more effective than existing models that do not explicitly account for temporal information~\citep{music:turnbull08,mandel08,eck08,hoffman09}.%

\subsubsection{Music dataset}
We consider the CAL500 collection from \citep{music:turnbull08}, which consists of 502 songs 
and provides binary annotations with respect to a vocabulary ${\cal V}$ of 149 tags, ranging from genre and instrumentation, to mood and usage. 
To represent the acoustic content of a song we extract a time series of audio features ${\cal Y} = \{y_1, \dots, y_{|{\cal Y}|}\}$, by computing  the first 13 Mel frequency cepstral coefficients (MFCCs) \citep{Rabiner93} over half-overlapping windows of $46$ms of audio signal, augmented with first and second instantaneous derivatives. 
\ted{The song is then represented as a collection of 
 {\em audio fragments}, which are sequences of $T = 125$ audio features (approximately 6 seconds of audio), using a dense sampling with $80\%$ overlap.}

\subsubsection{Music annotation models}
Automatic music tagging is formulated as a supervised multi-label 
problem \citep{carneiro07},
where each class is a tag from ${\cal V}$. We approach this problem by modeling the audio content for each tag
with a H3M probability distribution. I.e., for each tag, we estimate an H3M over the audio fragments 
of the songs in the database that have been associated with that tag, using 
the hierarchical estimation procedure based on VHEM-H3M.
More specifically, the database is first processed at the song level, using the EM algorithm to learn a H3M with $K\s{s} = 6$ components for each song\footnote{Most pop songs have $6$ structural parts: intro, verse, chorus, solo, bridge and outro.} from its audio fragments.
Then, for each tag, the song-level H3Ms labeled with that tag 
are pooled together to form a large H3M, and the VHEM-H3M algorithm is used to reduce this to a
final H3M tag-model with $K = 3$ components ($\tau = 10$ and $N = N_v N_t K\s{s}$, where $N_v = 1000$ and $N_t$ is the number of training songs for the particular tag).

Given the tag-level models, a song can be represented as a vector of posterior probabilities of each tag (a semantic multinomial, SMN), by extracting features from the song, computing the likelihood of the features under each tag-level model, and applying Bayes' rule. 
A test song is annotated with the top-ranking tags in its SMN.
To retrieve songs given a tag query, a collection of songs is ranked by the tag's probability in their SMNs.

We compare VHEM-H3M with three alternative algorithms for estimating the H3M tag models: PPK-SC, PPK-SC-hybrid, and EM-H3M.\footnote{For this experiment, we were not able to successfully estimate accurate H3M tag models with SHEM-H3M.  
In particular, SHEM-H3M requires generating an appropriately large number of real samples to produce accurate estimates. However, due to the computational limits of our cluster, we were able to test SHEM-H3M only using a small number of samples. In preliminary experiments we registered performance only slightly above chance level and training times still twice longer than for VHEM-H3M.
For a comparison between VHEM-H3M and SHEM-H3M on density estimation, the reader can refer to the experiment in Section \ref{sec:exp:dig} on online hand-writing classification and retrieval.}
\ted{For all three alternatives, we use the same number of mixture components in the tag models ($K=3$).}
%
%
For the two PPK-SC methods, we leverage
the work of \cite{jebara2007} to learn H3M tag models, and use it in place of the 
VHEM-H3M algorithm in the second stage of the hierarchical estimation procedure. 
We found that it was necessary to implement the PPK-SC approaches with song-level H3Ms with only $K\s{s} = 1$ component (i.e., a single HMM),  since the computational cost for constructing the initial embedding scales poorly with the number of input HMMs.\footnote{Running PPK-SC with $K\s{s} = 2$ took 3958 hours in total (about 4 times more than when setting $K\s{s} = 1$), with no improvement in annotation and retrieval performance. A larger $K\s{s}$ would yield impractically long learning times.}
PPK-SC first 
applies spectral clustering to the song-level HMMs 
and then selects as the cluster centers the HMMs that map closest to the spectral cluster centers in the spectral embedding. 
PPK-SC-hybrid is a hybrid method combining PPK-SC for clustering, and VHEM-H3M for estimating the cluster centers.  
Specifically, 
after spectral clustering, HMM
cluster centers are estimated by applying VHEM-H3M  (with $K\s{r}=1$) to the HMMs assigned to each of the resulting clusters. 
In other words, PPK-SC and PPK-SC-hybrid use spectral clustering to summarize a collection of song-level HMMs with a few HMM centers, forming a H3M tag model.
The mixture weight of each HMM component (in the H3M tag model) is set proportional to the number of HMMs assigned to that cluster.
%
 %

For EM-H3M, the H3M tag models were estimated directly from 
the audio fragments from the relevant songs using the EM-H3M algorithm.\footnote{The EM algorithm has been used to estimate HMMs from music data in previous work, e.g., \citep{scaringella2005modeling,reed06}.}
%
Empirically, we found that, due to its runtime and RAM requirements, for EM-H3M we must use non-overlapping audio-fragments and evenly subsample by $73\%$ on average, resulting in $14.5\%$ of the sequences used by VHEM-H3M. Note that, however, EM-H3M is still using $73\%$ of the actual song data (just with non-overlapping sequences). We believe this to be a reasonable comparison between EM and VHEM, as both methods use roughly similar resources (the sub-sampled EM is still $3$ times slower, as reported in Table \ref{tab:cal500}). Based on our projections, running EM over densely sampled song data would require roughly $9000$ hours of CPU time (e.g., more than $5$ weeks when parallelizing the algorithm over $10$ processors), as opposed to 630 hours for VHEM-H3M. This would be extremely cpu-intensive given the computational limits of our cluster.
The VHEM algorithm, on the other hand, can learn from considerable amounts of data while still maintaining low runtime and memory requirements.%
\footnote{For example, consider learning a tag-level H3M from $200$ songs, which corresponds to over $3$GB of audio fragments. 
Using the hierarchical estimation procedure, we first model each song (in average, $15$MB of audio fragments) individually as a song-level H3M, and we save the song models (150 KB of memory each). Then, we pool the $200$ song models into a large H3M (in total 30MB of memory), and reduce it to a smaller tag-level H3M using the VHEM-H3M algorithm.}

\ted{Finally, we also compare against two state-of-the-art models for music tagging, 
HEM-DTM \citep{coviello2011}, which is based on a different time-series model (mixture of dynamic textures), and HEM-GMM \citep{music:turnbull08}, which is a bag-of-features model using GMMs.
Both methods use efficient hierarchical estimation based on a HEM algorithm \citep{Chan2010cvpr,Vasc1998} to obtain the tag-level models.\footnote{Both auto-taggers operate on audio features extracted over half-overlapping windows of $46$ms. HEM-GMM uses MFCCs with first and second instantaneous derivatives \citep{music:turnbull08}. HEM-DTM uses $34$-bins of Mel-spectral features  \citep{coviello2011}, which are further grouped in audio fragments of $125$ consecutive features.}}


\subsubsection{Performance metrics}

A test song is annotated with the 10 most likely tags, and annotation performance is measured with the per-tag precision (P), recall (R), and F-score (F), averaged over all tags.
If $|w_H|$ is the number of test songs that have the tag  $w$ in their ground truth annotations, $|w_A|$ is the number of times an annotation system uses $w$ when automatically tagging a song, and $|w_C|$ is the number of times $w$ is correctly used,
then precision, recall and F-score for the tag $w$ are defined as:
\begin{eqnarray}
\mbox{P} = \frac{|w_C|}{|w_A|}, \,\,\, \mbox{R} = \frac{|w_C|}{|w_H|}, \,\,\, \mbox{F} = 2\left( (\mbox{P})^{-1}+ (\mbox{R})^{-1} \right)^{-1}.
\label{eqn:prf}
\end{eqnarray}
Retrieval is measured 
by computing per-tag mean average precision (MAP) 
and precision at the first $k$ retrieved songs (P@$k$), for $k \in \{5,10,15\}$. 
The P@$k$ is the fraction of true positives in the top-$k$ of the ranking. MAP averages the precision at each point in the ranking where a song is correctly retrieved.
All reported metrics are averages over the 97 tags that have at least 30 examples in CAL500 (11 genre, 14 instrument, 25 acoustic quality, 6 vocal characteristics, 35 emotion and 6 usage tags), and are the result of 5-fold cross-validation.

Finally, we also record the total time (in hours) to learn the $97$ tag-level H3Ms on the 5 splits of the data.
For hierarchical estimation methods (VHEM-H3M and the PPK-SC approaches), this also includes the time to learn the song-level H3Ms.

		\begin{table}[t]
	\caption{Annotation and retrieval performance on CAL500, for VHEM-H3M, PPK-SC, PPK-SC-\emph{hybrid}, EM-H3M, HEM-DTM~\citep{coviello2011} and HEM-GMM~\citep{music:turnbull08}.}
		\label{tab:cal500}
	\begin{center}
	\begin{tabular}{lcccccccc}
	& \multicolumn{3}{c}{\bf annotation}  &\multicolumn{4}{c}{\bf retrieval}
\vspace{0.02in}\\
	& P & R & F & MAP  & P@5  &P@10  &P@15     & time ($h$)\\
	\hline \vspace{-.1in}\\
VHEM-H3M   			& 0.446	&0.211	&0.260	&0.440			&0.474	&0.451	&0.435	 & 629.5 \\ 
\hline
PPK-SC 	      			& 0.299	&0.159	&0.151	&0.347			&0.358	&0.340	&0.329	 &  974.0 \\	  
PPK-SC-\emph{hybrid} 	      			& 0.407	&0.200	&0.221	&0.415			&0.439	&0.421	&0.407	 & 991.7 \\	  
\hline
EM-H3M	      			&0.415	&0.214	&0.248	&0.423			&0.440	&0.422	&0.407	 & 1860.4 \\      
\hline
HEM-DTM      
		      			&0.431	&0.202	&0.252	&0.439			&0.479	&0.454	&0.428	 & - \\
HEM-GMM     			&0.374	&0.205	&0.213	&0.417			&0.441	&0.425	&0.416	 & -  \\
	\end{tabular}
	\end{center}
	\vspace{-.2in}
	\end{table}

\subsubsection{Results}
In Table \ref{tab:cal500} we report the performance of the various algorithms for both annotation and retrieval on the CAL500 dataset.  
Looking at the overall runtime, 
VHEM-H3M is the most efficient algorithm for estimating H3M distributions from music data, as it requires only $34\%$ of the runtime of EM-H3M, and $65\%$ of the runtime of PPK-SC.
The VHEM-H3M algorithm capitalizes on the first stage of song-level H3M estimation (about one third of the total time) by efficiently and effectively using the song-level H3Ms to learn the final tag models. 
Note that the runtime of PPK-SC corresponds to setting $K\s{s} =1$. When we set $K\s{s} = 2$, we registered a running time four times longer, with no significant improvement in performance.

The gain in computational efficiency does not negatively affect the quality of the corresponding models.
On the contrary, VHEM-H3M achieves better performance than EM-H3M,\footnote{The differences in performance are statistically significant based on a paired t-test with $95\%$ confidence.} strongly improving the top of the ranked lists, as evinced by the higher P@$k$ scores. 
Relative to EM-H3M, VHEM-H3M has the benefit of regularization, and during learning can efficiently leverage all the 
music data condensed in the song H3Ms.
VHEM-H3M also outperforms both PPK-SC approaches on all metrics.
PPK-SC discards considerable information on the clusters' structure by selecting one of the original HMMs to approximate each cluster. 
This significantly affects the accuracy of the resulting annotation models.
VHEM-H3M, on the other hand, generates novel HMM cluster centers to summarize the clusters. This allows to retain more accurate information in the final annotation models.

PPK-SC-hybrid achieves considerable improvements relative to standard PPK-SC, at relatively low additional computational costs.\footnote{In PPK-SC-hybrid, each run of the VHEM-H3M algorithm converges quickly since there is only one HMM component to be learned, and can benefit from clever initialization (i.e., to the HMM mapped the closest to the spectral clustering center).}  
This further demonstrates that the VHEM-H3M algorithm can effectively summarize in a smaller model the information contained in \emph{several} HMMs.
In addition, we observe that VHEM-H3M still outperforms PPK-SC-hybrid, suggesting that the former produces more accurate \ted{cluster centers and density estimates}. 
In fact, VHEM-H3M \emph{couples} clustering and learning HMM cluster centers, and is entirely based on maximum likelihood for estimating the H3M annotation models. PPK-SC-hybrid, on the contrary, separates clustering and parameter estimation, and optimizes them against two different metrics (i.e., PPK similarity and expected log-likelihood, respectively).
As a consequence, the two phases may be mismatched, and the centers learned with VHEM may not be the best representatives of the clusters according to PPK affinity.


Finally, VHEM-H3M compares favorably to the auto-taggers based on other generative models. First, VHEM-H3M outperforms HEM-GMM, which does not model temporal information in the audio signal, on all metrics.
Second, the performances of VHEM-H3M and HEM-DTM (a continuous-state temporal model) are not statistically different based on a paired t-test with $95\%$ confidence, except for annotation precision where VHEM-H3M scores significantly higher.
{Since HEM-DTM is based on linear dynamic systems (a continuous-state model), it can model stationary time-series in a linear subspace. In contrast, VHEM-H3M uses HMMs with discrete states and GMM emissions, and can hence better adapt to non-stationary time-series on a non-linear manifold. This difference is illustrated in the experiments: VHEM-H3M outperforms HEM-DTM on the human MoCap data (see Table \ref{tab:mocap}), which has non-linear dynamics, while the two perform similarly on the music data (see Table \ref{tab:cal500}), where audio features are often stationary over short time frames.


\subsection{On-line hand-writing data classification and retrieval}	
\label{sec:exp:dig}

In this experiment, we investigate the performance of the VHEM-H3M algorithm in estimating class-conditional H3M distributions for automatic classification and retrieval of on-line hand-writing data.

\subsubsection{Dataset}

We consider the Character Trajectories Data Set \citep{Frank+Asuncion:2010}, which is a collection of $2858$ examples of characters 
from the same writer, originally compiled to study handwriting motion primitives \citep{williams2006extracting}. Each example is the trajectory of one of the $20$ different characters that correspond to a single pen-down segment.
The data was captured from a WACOM tablet at 200 Hz, and consists of $(x,y)$-coordinates and pen tip force. The data has been numerically differentiated and Gaussian smoothed \citep{williams2006extracting}.
Half of the data is used for training, with the other half held out for testing. 

\subsubsection{Classification models and setup}

From the hand-writing examples in the training set, we estimate a series of class-conditional H3M distributions, one for each character, using hierarchical estimation with the VHEM-H3M algorithm. 
%
%
First, for each character, we partition all the relevant training data into groups of $3$ sequences, and learn a HMM (with $S=4$ states and $M=1$ component for the GMM emissions) from each group using the Baum-Welch algorithm. 
Next, we estimate the class-conditional distribution (classification model) for each character
 by aggregating all the relevant HMMs and summarizing them into a H3M with $K = 4$ components using the VHEM-H3M algorithm ($\tau=10$ and $N=N_v N_c$, where $N_v=10$,%
\footnote{Note that choosing a lower value of $N_v$ (compared to the music experiments) plays a role in making the clustering algorithm more \emph{reliable}.
Using fewer virtual samples equates to attaching smaller ``virtual probability masses'' to the input HMMs, and leads to \emph{less certain} assignments of the input HMMs to the clusters (cf. equation \ref{eqn:zijhat}).
This determines more mixing in the initial iterations of the algorithm (e.g., similar to higher annealing temperature), 
and reduces the risk
of prematurely specializing any cluster to one of the original HMMs.
This effect is desirable, since the input HMMs are estimated over a smaller number of sequences  (compared to the music experiments) and can therefore be noisier and less reliable. 
}
and $N_c$ is the number of intermediate models for that character).
Using the character-level H3Ms and Bayes' rule, for each hand-writing example in the test set we compute the posterior probabilities of all of the 20 characters.
Each example is classified as the character with largest posterior probability.
For retrieval given a query character, examples in the test set are ranked by the character's posterior probability.


We repeated the same experiment using PPK-SC or SHEM-H3M ($\tau=10$, $N=1000$) 
to estimate the classification 
models from the intermediate HMMs.
Finally, we considered the EM-H3M algorithm, which directly uses the training sequences to learn the class-conditional H3M ($K=4$).


Since VHEM-H3M, SHEM-H3M and EM-H3M are \emph{iterative} algorithms, we studied them when varying the stopping criterion. 
In particular, the algorithms were terminated when the relative variation in the value of the objective function between two consecutive iterations was lower than a threshold $\Delta LL $, which we varied in $\{10^{-2},10^{-3},10^{-4},10^{-5}\}$.\footnote{In a similar experiment where we used the number of iterations as the stopping criterion, we registered similar results.}


Finally, we measure classification and retrieval performance on the test set using the classification accuracy, and the average per-tag P@$3$ and P@$5$. 
We also report the total training time, which includes the time used to learn the intermediate HMMs. The experiments consisted of 5 trials with different random initializations of the algorithms.

\subsubsection{Results}

	\begin{table}[t]
	\caption{Classification and retrieval performance, and training time on the Character Trajectories Data Set, for VHEM-H3M, PPK-SC, SHEM-H3M, and EM-H3M.}
		\label{tab:ChTr}
	\begin{center}
	\begin{tabular}{cccccr}
	& stop    &\multicolumn{2}{c}{\bf retrieval} & {\bf classification} &
\vspace{0.02in}\\
& $\Delta LL$	    & P@3  &P@5   	&Accuracy      & total time (s)\\
	\hline \vspace{-.1in}\\
\multirow{5}{*}{VHEM-H3M} 	&$10^{-2}$&			0.750&	0.750&		0.618&	1838.34\\
						&$10^{-3}$&			0.717&	0.750&		0.619&	1967.55\\
						&$10^{-4}$&			0.733&	0.790&		0.648&	2210.77\\
						&$10^{-5}$&			0.750&	0.820&		0.651&	2310.93\\
\hline
\multirow{5}{*}{SHEM-H3M} 	&$10^{-2}$&			0.417&	0.440&		0.530&	13089.32\\
						&$10^{-3}$&			0.683&	0.680&		0.664&	23203.44\\
						&$10^{-4}$&			0.700&	0.750&		0.689&	35752.20\\
						&$10^{-5}$&			0.700&	0.750&		0.690&	50094.36\\
\hline
\multirow{5}{*}{EM-H3M} 		&$10^{-2}$&			0.583&	0.610&		0.646&	6118.53\\
						&$10^{-3}$&			0.617&	0.650&		0.674&	7318.56\\
						&$10^{-4}$&			0.650&	0.710&		0.707&	9655.08\\
						&$10^{-5}$&			0.517&	0.560&		0.635&	10957.38\\
\hline
PPK-SC							&      -	&	  	0.600&	0.700&	0.646&	1463.54 \\
	\end{tabular}
	\end{center}
	\vspace{-.2in}
	\end{table}


Table \ref{tab:ChTr} lists the classification and retrieval performance on the test set for the various methods.
%
%
Consistent with the experiments on music annotation and retrieval (Section \ref{sec:exp:mus}), VHEM-H3M performs better than PPK-SC on all metrics. By learning novel HMM cluster centers, VHEM-H3M estimates H3M distributions that are representative of all the relevant intermediate HMMs, and hence of all the relevant training sequences.
While EM-H3M is the best in classification (at the price of longer training times), VHEM-H3M performs better in retrieval, as evinced by the P@$3$ and P@$5$ scores. 
In terms of training time, VHEM-H3M and PPK-SC are about $5$ times faster than EM-H3M. In particular, PPK-SC is the fastest algorithm, since the small number of input HMMs (i.e., on average $23$ per character) allows to build the spectral clustering embedding efficiently.

The version of HEM based on actual sampling (SHEM-H3M) performs better than VHEM-H3M in classification, but VHEM-H3M has higher retrieval scores. 
However, the training time for SHEM-H3M is approximately $15$ times longer than for VHEM-H3M.
In order to reliably estimate the reduced models, the SHEM-H3M algorithm requires generating a large number of samples, and computing their likelihood at each iteration. 
In contrast, the VHEM-H3M algorithm efficiently approximates the sufficient statistics of a virtually unlimited number of samples, without the need of using real samples.

It is also interesting to note that EM-H3M appears to suffer from overfitting of the training set, as suggested by the {\em overall drop} in performance when the stopping criterion changes from $\Delta LL=10^{-4}$ to $\Delta LL=10^{-5}$.
In contrast, both VHEM-H3M and SHEM-H3M consistently improve on all metrics as the algorithm converges (again looking at $\Delta LL \in \{10^{-4}, 10^{-5}\}$).  
These results suggest that the regularization effect of hierarchical estimation, which is based on averaging over 
{\em more} samples (either virtual or actual),  can positively impact the generalization of the learned models.\footnote{For smaller values of $\Delta LL$ (e.g., $\Delta LL < 10^{-5}$), the performance of EM-H3M did not improve.}

Finally, we elaborate
on how these results compare to the experiments on music annotation and retrieval (in Section \ref{sec:exp:mus}).
First, in the Character Trajectory Data Set the number of training sequences associated with each class (i.e, each character) is small compared to the CAL500 dataset.\footnote{In the Character Trajectory dataset there are on average $71$ training sequences per character. In CAL500, each tag is associated with \emph{thousands} of training sequences at the song level 
(e.g., an average of about $8000$ audio fragments per tag).
}
As a result, the EM-H3M algorithm is able to process all the data, and achieve good classification performance.
However, EM-H3M still needs to evaluate the likelihood of all the original sequences at each iteration.  This leads to 
slower iterations, and results in a total training time about $5$ times longer than that of VHEM-H3M (see Table \ref{tab:ChTr}).
Second, the Character Trajectory  data is more ``controlled'' than the CAL500 data, since each class corresponds to a single character, and all the examples are from the same writer.
As a consequence, there is less variation in the intermediate HMMs (i.e., they are clustered more closely), and several of them may summarize the cluster well, providing good candidate cluster centers for PPK-SC.
In conclusion, PPK-SC faces only a limited loss of information when selecting one of the initial HMMs to represent each cluster, 
 and achieves reasonable performances.

\subsection{Robustness of VHEM-H3M to number and length of virtual samples} 
\label{sec:robustness}



	\begin{table}[t]
	\caption{Annotation and retrieval performance on CAL500 for VHEM-H3M when varying the virtual sample parameters $N_v$ and $\tau$. 
	 }
		\label{tab:rob_meta}
	\begin{center}
	\begin{tabular}{c|cccc|cccc}
	& \multicolumn{4}{c|}{ $N_v = 1000$}  &\multicolumn{4}{c}{ $\tau = 10$}\\
	& $\tau= 2$ & $\tau= 5$ & $\tau= 10$ & $\tau= 20$ & $N_v= 1$ & $N_v = 10$ & $N_v = 100$ & $N_v = 1000$ \\
	\hline 
	P@5				&0.4656   & 0.4718  &  0.4738  &  0.4734 & 0.4775  &  0.4689   & 0.4689    &0.4738 \\
	P@10			&0.4437   & 0.4487   & 0.4507  &  0.4478 & 0.4534  &  0.4491   &  0.4466   & 0.4507 \\
	P@15			&0.4236   & 0.4309  &  0.4346  &  0.4327  & 0.4313  &  0.4307   & 0.4242    &0.4346 \\
	\end{tabular}
	\end{center}
	\vspace{-.2in}
	\end{table}


The generation of virtual samples in VHEM-H3M is controlled by two parameters:
the number of virtual sequences ($N$), and their length ($\tau$). 
In this section, we investigate the impact of these parameters on annotation and retrieval performance on CAL500.
For a given tag $t$, we set $N = N_v N_t K\s{s}$, where $N_v$ is a constant, $N_t$ the number of training songs for the tag, and $K\s{s}$ the number of mixture components for each song-level H3M.
Starting with $(N_v,\tau) = (1000,10)$, each parameter is varied while keeping the other one fixed, and annotation and retrieval performance on the CAL500 dataset are calculated, as described in Section \ref{sec:exp:mus}. 

Table \ref{tab:rob_meta} presents the results, for $\tau \in \{2,5,10,20\}$ and $N_v  \in \{1,10,100,1000\}$.
The performances when varying $\tau$ are close on all metrics. For example, average P@5, P@10 and P@15 vary in small ranges ($0.0082$, $0.0070$ and $0.0110$, respectively).
Similarly, varying the number of virtual sequences $N_v$ has a limited impact on performance as well.\footnote{Note that the E-step of the VHEM-H3M algorithm averages over all possible observations compatible with the input models, also when we choose a low value of  $N_v$ (e.g., $N_v=1$). The number of virtual samples controls the ``virtual mass'' of each input HMMs and thus the certainty of cluster assignments.}
This demonstrates that VHEM-H3M is fairly robust to the choice of these parameters. 

Finally, we tested VHEM-H3M for music annotation and retrieval on CAL500, using virtual sequences of the same length as 
the audio fragments used at the song level,
i.e., $\tau = T = 125$.
Compared to $\tau = 10$ (the setting used in earlier experiments), we registered an $84\%$ increase in total running time, with no corresponding improvement in performance. 
Thus, \ted{in our experimental setting}, making the virtual sequences relatively short positively impacts the running time, without reducing the quality of the learned models.

%



\section{Conclusion}

In this paper, we presented a variational HEM (VHEM) algorithm for clustering HMMs with respect to their probability distributions, while generating a novel HMM center to represent each cluster.
Experimental results demonstrate the efficacy of the algorithm on various clustering, annotation, and retrieval problems involving time-series data, showing improvement over current methods.
In particular, using a two-stage, \emph{hierarchical estimation} procedure --- learn H3Ms on many smaller subsets of the data, and summarize them in a more compact H3M model of the data  --- the VHEM-H3M algorithm estimates annotation models from 
data 
more efficiently than standard EM and also improves model robustness through better regularization. Specifically, 
averaging over all possible virtual samples prevents over-fitting, which can improve the generalization of the learned models.
Moreover, using relatively short virtual sequences positively impacts the running time of the algorithm, without negatively affecting its performance on practical applications.
In addition, we have noted that the VHEM-H3M algorithm is robust to the choice of the number and length of virtual samples. 

In our experiments, we have implemented the first stage of the hierarchical estimation procedure by partitioning data in \emph{non-overlapping} subsets (and learning an intermediate H3M on each subset).
In particular, partitioning the CAL500 data at the song level had a practical advantage.
Since individual songs in CAL500 
are relevant to several tags, the estimation of the song H3Ms can be executed one single time for each song in the database, and \emph{re-used} in the VHEM estimation of all the associated tag models. This has a positive impact on computational efficiency.
Depending on the particular application, however, a slightly different implementation of this first stage (of the hierarchical estimation procedure) may be better suited.
For example, when estimating a H3M from a very large amount of training data, one could use a procedure that does not necessarily cover all data,
inspired by \cite{kleinerbootstrapping}. 
If $n$ is the size of the training data,
first estimate $B>1$ intermediate H3Ms on as many (possibly overlapping) ``little'' bootstrap subsamples of the data,\footnote{Several techniques have been proposed to bootstrap from sequences of samples, for example \citep{hall1995blocking}.} each of size $b<n$. 
Then summarize all the intermediate H3Ms into a final H3M using the VHEM-H3M algorithm.


In future work we plan to extend VHEM-H3M to the case where all HMMs share a large GMM universal background model for the emission distributions (with each HMM state having a different set of weights for the Gaussian components), which is commonly used in speech \citep{huang1989semi,bellegarda1990tied,Rabiner93}.
This would allow for faster training (moving the complexity to estimating the noise background model) and would require a faster implementation of the inference (e.g., using a strategy similar to \citep{coviello2012growing}). 
In addition, we plan to derive a HEM algorithm for HMMs with discrete emission distributions, and compare its performance to the work presented here and to the extension with the large GMM background model.

%



\acks E.C. thanks 
Yingbo Song for providing the code of \citep{jebara2007} and assistance with the experiments on motion clustering. The authors acknowledge support from Google, Inc. E.C. and G.R.G.L. also wish to acknowledge support from Qualcomm, Inc., Yahoo! Inc., the Hellman Fellowship Program, and the National Science Foundation (grants CCF-0830535 and IIS-1054960). A.B.C. was supported by the Research Grants Council of the Hong Kong Special Administrative Region, China (CityU 110610). G.R.G.L. acknowledges support from the Alfred P. Sloan Foundation. This research was supported in part by the UCSD FWGrid Project, NSF Research Infrastructure Grant Number~EIA-0303622.

\appendix

\section*{Appendix A. Derivation of the E-step}
\label{app:e_step}

\noindent

%

The maximization of \refeqn{eqn:Lhmm} with respect to $\phi^{i,j}_{t}(\rho_t|\rho_{t-1},\beta_t)$ and $\phi^{i,j}_{1}(\rho_1|\beta_1)$ is carried out independently for each pair $(i,j)$, and follow \citep{hershey2008variational}.
In particular it uses a backward recursion,
starting with $\calL^{i,j}_{\tau+1}(\beta_{t},\rho_{t}) = 0$, for $t = \tau,\dots,2$, 
	\begin{align}
	\hat\phi^{i,j}_{t}(\rho_t|\rho_{t-1},\beta_t) = { } & \frac{\st{r}{j}{\rho}{t-1}{t} \exp \left\{ \calL\subgm^{(i,\beta_t),(j,\rho_t)}+ \calL^{i,j}_{t+1}(\beta_{t},\rho_{t}) \right\}}{\sum_{\rho}^{S} \st{r}{j}{\rho}{t-1}{} \exp \left\{ \calL\subgm^{(i,\beta_t),(j,\rho)} + \calL^{i,j}_{t+1}(\beta_{t},\rho) \right\}} \\
         \calL^{i,j}_t(\beta_{t-1},\rho_{t-1})            = { } & \sum_{\beta =1}^{S} \st{b}{i}{\beta}{t-1}{} \log \sum_{\rho =1}^{S} \st{r}{j}{\rho}{t-1}{} \exp \left\{ \calL\subgm^{(i,\beta),(j,\rho)}  + \calL^{i,j}_{t+1}(\beta,\rho) \right\}
         ,
	\end{align}
 and terminates with
	\begin{align}
         \hat\phi^{i,j}_{1}(\rho_1|\beta_1) = { } & \frac{\is{r}{j}{\rho}{1} \exp \left\{ \calL\subgm^{(i,\beta_1),(j,\rho_1)} + \calL^{i,j}_{2}(\beta_{1},\rho_{1}) \right\}}{\sum_{\rho}^{S} \is{r}{j}{\rho}{{}} \exp \left\{ \calL\subgm^{(i,\beta_1),(j,\rho)} + \calL^{i,j}_{2}(\beta_{1},\rho) \right\}} \\
        \calL \subhmm ^{i,j} 
        = { } & \sum_{\beta =1}^{S} \is{b}{i}{\beta}{{}} \log \sum_{\rho =1}^{S} \is{r}{j}{\rho}{{}} \exp \left\{ \calL\subgm^{(i,\beta),(j,\rho)} + \calL^{i,j}_{2}(\beta,\rho) \right\} \label{eqn:opt_el_hmm_1}
	\end{align}
where \refeqn{eqn:opt_el_hmm_1} is the maxima of the terms in \refeqn{eqn:Lhmm} in Section \ref{text:VD}.

\section*{Appendix B. Derivation of the M-step}
\label{app:m_step}

\noindent

The M-steps involves maximizing  the lower bound in \refeqn{eqn:LB-cost-summary} with respect to ${\calM\s{r}} $, while holding the variational distributions fixed, 
\begin{align}
		{\calM\s{r}}^{*}  =   \argmax_{ \calM\s{r}}  \sum_{i=1}^{K\s{b}}  \calL_{H3M}^i.
	\label{eqn:Mstep}
\end{align}
Substituting \refeqn{eqn:LB-H3M-final} and \refeqn{eqn:LB-HMM-final} into the objective function of \refeqn{eqn:Mstep},
	\begin{align}
	\lefteqn{\calL(\calM\s{r}) =  \sum_{i=1}^{K\s{b}}  \calL_{H3M}^i} \\
	&=
	  \sum_{i,j} 
	    \hat{z}_{ij} \left\{ \log \frac{\omega\s{r}_j}{\hat{z}_{ij}} 
	 +  N_i
	 	\sum_{\beta_{1:\tau}} \pi_{\beta_{1:\tau}}^{(b),i}
	\sum_{\rho_{1:\tau}} 
	\hat{\phi}^{i,j}(\rho_{1:\tau}|\beta_{1:\tau}) 
	\left[
	\log \frac{\pi^{(r),j}_{\rho_{1:\tau}}}{\hat{\phi}^{i,j}(\rho_{1:\tau}|\beta_{1:\tau}) }
	+ \sum_t \calL_{GMM}^{(i,\beta_t),(j,\rho_t)}
	\right]
	 \right\} 
	 \label{eqn:M-step-objective}
	\end{align}
In the following, we detail the update rules for the parameters of the reduced model $\calM\s{r}$.

\subsection*{HMMs mixture weights}

Collecting terms in \refeqn{eqn:M-step-objective} that only depend on the mixture weights $\{\omega\s{r}_j\}_{j=1}^{K\s{r}}$, we have
	\begin{align}
	\widetilde{\calL}(\{\omega\s{r}_j\}) =  \sum_{i} 
	    \sum_{j} \hat{z}_{ij} \log \omega\s{r}_j
	    = \sum_{j} \left[\sum_i \hat{z}_{ij}\right] \log \omega\s{r}_j
	    \label{eqn:obj-weights}
	\end{align}
Given the constraints $\sum_{j=1}^{K\s{r}} \omega\s{r}_j=1$ and $\omega\s{r}_j\geq0$, 
\refeqn{eqn:obj-weights} is maximized using the result in Appendix C.1, which yields the update  in 
\refeqn{eqn:M-step-omega}.


\subsection*{Initial state probabilities}

The objective function in \refeqn{eqn:M-step-objective} factorizes for each HMM $\calM\s{r}_j$, and hence the parameters of each HMM are updated independently.
For the $j$-th HMM, we collect terms in \refeqn{eqn:M-step-objective} that depend on the initial state probabilities $\{\is{r}{j}{\rho}{{}}\}_{\rho=1}^{S}$,
	\begin{align}
	\widetilde{\calL}_j(\{\is{r}{j}{\rho}{{}}\}) &= 
	\sum_{i} \hat{z}_{ij} 
		N_i \sum_{\beta_1} \pi_{\beta_{1}}^{(b),i}
	\sum_{\rho_{1}} 
	\hat{\phi}^{i,j}_{1}(\rho_{1}|\beta_{1}) 
	\log \pi^{(r),j}_{\rho_{1}}
	\\
	&=
	\sum_{\rho_{1}} 
	\sum_{i} \hat{z}_{ij} 
	N_i \underbrace{
	\sum_{\beta_1} \pi_{\beta_{1}}^{(b),i}
	\hat{\phi}^{i,j}_{1}(\rho_{1}|\beta_{1}) 
	}_{\hat{\nu}^{i,j}_1(\rho_1)}
	\log \pi^{(r),j}_{\rho_{1}}
	\\
	\label{eqn:obj-pi-deriv}
	&=
	\sum_{\rho} 
	\sum_{i} \hat{z}_{ij} 
	N_i \hat{\nu}^{i,j}_1(\rho)
	\log \pi^{(r),j}_{\rho}
	\\
	&\propto 
	\sum_{\rho} 
	\left[\sum_{i} \hat{z}_{ij} 
	\omega\s{b}_i \hat{\nu}^{i,j}_1(\rho)
	\right]
	\log \pi^{(r),j}_{\rho}
	,
	\label{eqn:obj_f_pirj}
	\end{align}
where in the \refeqn{eqn:obj-pi-deriv} we have used the summary statistic defined in \refeqn{eq:sum_stat1}.
%
%
Considering the constraints $\sum_{\rho=1}^{S} \is{r}{j}{\rho}{{}} =1$ and $\is{r}{j}{\rho}{{}} \geq 0$, 
\refeqn{eqn:obj_f_pirj} is maximized using the result in Appendix C.1, giving the update formula in
\refeqn{eqn:M-step-pi-a}.


\subsection*{State transition probabilities}

Similarly, for each HMM $\calM\s{r}_j$ and previous state $\rho$, we collect terms in \refeqn{eqn:M-step-objective} that depend on the transition probabilities $\{a^{(r),j}_{\rho,\rho'}\}_{\rho'=1}^{S}$,
	\begin{align}
	\lefteqn{
	\widetilde{\calL}_{j,\rho}(\{a^{(r),j}_{\rho,\rho'}\}_{\rho'=1}^{S}) 
	=
	  \sum_{i} 
	    \hat{z}_{ij} N_i
	 	\sum_{\beta_{1:\tau}} \pi_{\beta_{1:\tau}}^{(b),i}
	\sum_{\rho_{1:\tau}} 
	\hat{\phi}^{i,j}(\rho_{1:\tau}|\beta_{1:\tau}) 
	\log \pi^{(r),j}_{\rho_{1:\tau}}
	}
	\\
	&\propto
	 	  \sum_{i} 
	    \hat{z}_{ij} N_i
	 	\sum_{\beta_{1:\tau}} \left[\pi^{(b),i}_{\beta_1} \prod_{t=2}^{\tau} a^{(b),i}_{\beta_{t-1},\beta_t}\right]
		\sum_{\rho_{1:\tau}} \left[\hat{\phi}^{i,j}_1(\rho_{1}|\beta_{1}) \prod_{t=2}^{\tau}	\hat{\phi}^{i,j}_t(\rho_{t}|\rho_{t-1},\beta_{t}) \right]
		\left[\sum_{t=2}^{\tau}\log a^{(r),j}_{\rho_{t-1},\rho_t}\right]
	\\
	&= 	 	  \sum_{i}  \hat{z}_{ij} N_i
	\sum_{\rho_1} 	\sum_{\rho_2} 	\sum_{\beta_2} 
	\underbrace{\sum_{\beta_1} 
	\underbrace{ \pi^{(b),i}_{\beta_1} \hat{\phi}^{i,j}_1(\rho_{1}|\beta_{1}) 
	}_{\nu_1^{i,j}(\rho_1,\beta_1) } a^{(b),i}_{\beta_1,\beta_2} \hat{\phi}^{i,j}_2(\rho_{2}|\rho_{1},\beta_{2})
	}_{\xi_2^{i,j}(\rho_1,\rho_2,\beta_2)}
		\left[\log a^{(r),j}_{\rho_{1},\rho_2} \right.
	\\
	& \qquad	
	 + \left. \sum_{\beta_{3:\tau}}  \sum_{\rho_{3:\tau}} 
		  \prod_{t=3}^{\tau} a^{(b),i}_{\beta_{t-1},\beta_t}
		\prod_{t=3}^{\tau}	\hat{\phi}^{i,j}_t(\rho_{t}|\rho_{t-1},\beta_{t})
		 \sum_{t=3}^{\tau}\log a^{(r),j}_{\rho_{t-1},\rho_t} \right]
	\\
		&= 	 	  \sum_{i}  \hat{z}_{ij} N_i
	\sum_{\rho_1} 	\sum_{\rho_2} 	\sum_{\beta_2} 
	{\xi_2^{i,j}(\rho_1,\rho_2,\beta_2)} 
		\log a^{(r),j}_{\rho_{1},\rho_2}
		\\
	& \quad +  \sum_{i}  \hat{z}_{ij} N_i
 	\sum_{\rho_2}  	\sum_{\rho_3} 	 \sum_{\beta_3} 
	\underbrace{
	\sum_{\beta_2} 
	\underbrace{
		\sum_{\rho_1}
	{\xi_2^{i,j}(\rho_1,\rho_2,\beta_2)}
	}
	_{\nu_2^{i,j}(\rho_2,\beta_2)}  a^{(b),i}_{\beta_2,\beta_3} \hat{\phi}^{i,j}_3(\rho_{3}|\rho_{2},\beta_{3})
	}_{\xi_3^{i,j}(\rho_2,\rho_3,\beta_3)}
	\left[ \log a^{(r),j}_{\rho_{2},\rho_3} \right.
	\\
	& \qquad + \left.
	 \sum_{\beta_{4:\tau}}  \sum_{\rho_{4:\tau}} 
	 \prod_{t=4}^{\tau} a^{(b),i}_{\beta_{t-1},\beta_t}
		\prod_{t=4}^{\tau}	\hat{\phi}^{i,j}_t(\rho_{t}|\rho_{t-1},\beta_{t})
		 \sum_{t=4}^{\tau}\log a^{(r),j}_{\rho_{t-1},\rho_t} \right]
				\\
	& = \dots\\
	&=   \sum_{i}  \hat{z}_{ij} N_i
	\sum_{t=2}^{\tau}	\sum_{\rho_{t-1}} 	\sum_{\rho_t} 	\sum_{\beta_t} 
	{\xi_t^{i,j}(\rho_{t-1},\rho_t,\beta_t)} 
		\log a^{(r),j}_{\rho_{t-1},\rho_t}
	\\
	& \propto  \sum_{i}  \hat{z}_{ij} N_i
 	\sum_{\rho'} 	
	\underbrace{
	\sum_{t=2}^{\tau}\sum_{\beta} 
	{\xi_t^{i,j}(\rho,\rho',\beta)} 
	}_{\hat{\xi}^{i,j}(\rho,\rho')} 
		\log a^{(r),j}_{\rho,\rho'}
	\\
	&\propto  \sum_{\rho'=1}^S 
	\left[\sum_i \hat z_{ij} \omega\s{b}_i \hat\xi^{i,j}(\rho,\rho')\right]
	 \log a^{(r),j}_{\rho,\rho'}
	 .
	\label{eqn:obj_f_arj_rho}
	\end{align}
Considering the constraints $\sum_{\rho'=1}^{S}  a^{(r),j}_{\rho,\rho'}=1$ and $a^{(r),j}_{\rho,\rho'} \geq 0$, 
\refeqn{eqn:obj_f_arj_rho} is maximized using  the result in Appendix C.1, giving the update in 
\refeqn{eqn:M-step-pi-a}.



\subsection*{Emission probability density functions}
The cost function \refeqn{eqn:M-step-objective} factors also for each GMM indexed by $(j,\rho,\ell)$.
Factoring \refeqn{eqn:M-step-objective}, 
	\begin{align}
	\lefteqn{
	\widetilde{\calL}(\calM\s{r}_{j,\rho,\ell}) =
	  \sum_{i} 
	    \hat{z}_{ij}  N_i
	 	\sum_{\beta_{1:\tau}} \pi_{\beta_{1:\tau}}^{(b),i}
	\sum_{\rho_{1:\tau}} 
	\hat{\phi}^{i,j}(\rho_{1:\tau}|\beta_{1:\tau}) 
	\sum_t \calL_{GMM}^{(i,\beta_t),(j,\rho_t)}
	}
	\\
	&= \sum_{i} 
	    \hat{z}_{ij}  N_i
	 	\sum_{\beta_{1:\tau}} \left[ \pi_{\beta_{1}}^{(b),i} \prod_{t=2}^{\tau} a_{\beta_{t-1},\beta_t}^{(b),i} \right]
	\sum_{\rho_{1:\tau}} 
	\left[ \hat{\phi}_1^{i,j}(\rho_1|\beta_1)  \prod_{t=2}^{\tau}  \hat{\phi}_t^{i,j}(\rho_t|\rho_{t-1},\beta_t)  	\right]
	\sum_t \calL_{GMM}^{(i,\beta_t),(j,\rho_t)}
	\\
	&= \sum_{i} 
	    \hat{z}_{ij}  N_i
	 	\sum_{\rho_1}  
	\sum_{\beta_1} 
	\underbrace{
	\pi_{\beta_{1}}^{(b),i}
	\hat{\phi}^{i,j}_{1}(\rho_{1}|\beta_{1}) 
	}_{\nu^{i,j}_1(\rho_1,\beta_1)}
		 \left[ \calL_{GMM}^{(i,\beta_1),(j,\rho_1)}  \dots  \right.\\ 
	&\qquad \qquad \qquad  \qquad \qquad
  	\left.
		+ \sum_{\beta_{2:\tau}}
		 \prod_{t=2}^{\tau} a_{\beta_{t-1},\beta_t}^{(b),i}
	\sum_{\rho_{2:\tau}} 
	 \prod_{t=2}^{\tau}  \hat{\phi}^{i,j}(\rho_t|\rho_{t-1},\beta_t)  	
	\sum_{t=2}^{\tau} \calL_{GMM}^{(i,\beta_t),(j,\rho_t)} \right]
	\\
	& =  \sum_{i} 
	    \hat{z}_{ij}  N_i
\sum_{\rho_1} 			\sum_{\beta_1} 	
	\nu^{i,j}_1(\rho_1,\beta_1)
		\calL_{GMM}^{(i,\beta_1),(j,\rho_1)} \\
	&\quad+ \sum_i \hat{z}_{ij}  N_i
		\sum_{\rho_2} 		\sum_{\beta_2}
	\underbrace{
	\sum_{\rho_1} 		
	\underbrace{
	\sum_{\beta_1}  \left( 
	\nu^{i,j}_1(\rho_1,\beta_1)  a_{\beta_{1},\beta_2}^{(b),i}
	\right) \hat{\phi}_2^{i,j}(\rho_2|\rho_{1},\beta_2)  
	}_{\xi^{i,j}_2(\rho_1,\rho_2,\beta_2)}
	}_{\nu^{i,j}_2(\rho_2,\beta_2)}
	\left[  \calL_{GMM}^{(i,\beta_2),(j,\rho_2)}  \dots \right. \\
	& \qquad\qquad + \left.
		 \sum_{\beta_{3:\tau}}
		 \prod_{t=3}^{\tau} a_{\beta_{t-1},\beta_t}^{(b),i}
	\sum_{\rho_{3:\tau}} 
	 \prod_{t=3}^{\tau}  \hat{\phi}_t^{i,j}(\rho_t|\rho_{t-1},\beta_t)  	
	\sum_{t=3}^{\tau} \calL_{GMM}^{(i,\beta_t),(j,\rho_t)} \right]
	\\
	& = \dots \nonumber \\
	& = \sum_i  \hat{z}_{ij}  N_i 
	\sum_{t=1}^{\tau}
			\sum_{\rho_t} 	\sum_{\beta_t} 	
	\nu^{i,j}_t(\rho_t,\beta_t)
		\calL_{GMM}^{(i,\beta_t),(j,\rho_t)}  \\
	& \propto \sum_i  \hat{z}_{ij}  N_i 
			 	\sum_{\beta} 	
	\underbrace{
		\sum_{t=1}^{\tau} \nu^{i,j}_t(\rho,\beta)
		}_{\hat{\nu}^{i,j}_t(\rho,\beta)}
		\calL_{GMM}^{(i,\beta),(j,\rho)}  \\
	&\propto	
	  \sum_{i} 
	    \hat{z}_{ij}  N_i
	    \sum_{\beta=1}^S \hat{\nu}^{i,j}(\rho,\beta) 
	    		\sum_{m=1}^M c^{(b),i}_{\beta,m} 
	\,\,\hat{\eta}^{(i,\beta),(j,\rho)}_{\ell|m} \left[
	\log c^{(r),j}_{\rho,\ell} +
	\EV_{\calM\s{b}_{i,\beta,m}}[\log p(y|\calM\s{r}_{j,\rho,\ell})]
	\right]
	\\
	&=
	\Omega_{j,\rho}\left(
	\hat{\eta}^{(i,\beta),(j,\rho)}_{\ell|m} \left[
	\log c^{(r),j}_{\rho,\ell} +
	\EV_{\calM\s{b}_{i,\beta,m}}[\log p(y|\calM\s{r}_{j,\rho,\ell})]
	\right]
	\right)
	,
	\label{eqn:obj_f_emit_rj_rho}
	\end{align}
where in \refeqn{eqn:obj_f_emit_rj_rho} we use the weighted-sum operator defined in \refeqn{eqn:weightedsum}, which is over all base model GMMs $\{\calM\s{b}_{i, \beta, m}\}$.
The GMM mixture weights are subject to the constraints $\sum_{\ell = 1}^{M} c^{(r),j}_{\rho,\ell} = 1$, $\forall j, \rho$.
Taking the derivative with respect to each parameter and setting it to zero\footnote{We also considered the constraints on the covariance matrices $\Sigma^{(r),j}_{\rho,\ell} \succ \boldsymbol 0 $.}, 
gives the GMM update equations \refeqn{eq:updateTXT} and \refeqn{eq:updateSigmaTXT}.

\section*{Appendix C. Useful optimization problems}

\subsection*{Appendix C.1}
\label{sec:opt1}
The optimization problem
\begin{equation}
	\max_{\alpha_\ell} \quad   { } \sum_{\ell=1}^{L} \beta_\ell \log \alpha_\ell    	 \qquad
	\mbox{s.t.}	\qquad      { }  \sum_{\ell=1}^{L} \alpha_\ell = 1, \quad
					 { }  \alpha_\ell \geq 0,\, \forall \ell
\label{eqn:emax_form}
\end{equation}
is optimized by 
$	\alpha_\ell^* = \frac{\beta_\ell}{\sum_{\ell'=1}^{L} \beta_\ell'}$.

\subsection*{Appendix C.2}
\label{sec:opt2}
The optimization problem
\begin{equation}
	\max_{\alpha_\ell} \quad   { } \sum_{\ell=1}^{L} \alpha_\ell \left( \beta_\ell - \log \alpha_\ell  \right)  \qquad
	\mbox{s.t.}			\qquad   { } \sum_{\ell=1}^{L} \alpha_\ell = 1 \quad
					  { } \alpha_\ell \geq 0,\, \forall \ell
\label{eqn:gmax_form}
\end{equation}
is optimized by 
$	\alpha_\ell^* = \frac{\exp \beta_\ell}{\sum_{\ell'=1}^{L} \exp  \beta_\ell'}$.

\vskip 0.2in
\bibliography{References}

\begin{thebibliography}{47}
\providecommand{\natexlab}[1]{#1}
\providecommand{\url}[1]{\texttt{#1}}
\expandafter\ifx\csname urlstyle\endcsname\relax
  \providecommand{\doi}[1]{doi: #1}\else
  \providecommand{\doi}{doi: \begingroup \urlstyle{rm}\Url}\fi

\bibitem[Bahlmann and Burkhardt(2001)]{bahlmann2001measuring}
C.~Bahlmann and H.~Burkhardt.
\newblock {Measuring HMM similarity with the Bayes probability of error and its
  application to online handwriting recognition}.
\newblock In \emph{Document Analysis and Recognition, 2001. Proceedings. Sixth
  International Conference on}, pages 406--411. IEEE, 2001.

\bibitem[Banerjee et~al.(2005)Banerjee, Merugu, Dhillon, and
  Ghosh]{banerjee2005clustering}
A.~Banerjee, S.~Merugu, I.S. Dhillon, and J.~Ghosh.
\newblock {Clustering with Bregman divergences}.
\newblock \emph{The Journal of Machine Learning Research}, 6:\penalty0
  1705--1749, 2005.

\bibitem[Batlle et~al.(2002)Batlle, Masip, and Guaus]{batlle2002automatic}
E.~Batlle, J.~Masip, and E.~Guaus.
\newblock Automatic song identification in noisy broadcast audio.
\newblock In \emph{IASTED International Conference on Signal and Image
  Processing}. Citeseer, 2002.

\bibitem[Bellegarda and Nahamoo(1990)]{bellegarda1990tied}
J.R. Bellegarda and D.~Nahamoo.
\newblock {Tied mixture continuous parameter modeling for speech recognition}.
\newblock \emph{Acoustics, Speech and Signal Processing, IEEE Transactions on},
  38\penalty0 (12):\penalty0 2033--2045, 1990.

\bibitem[Carneiro et~al.(2007)Carneiro, Chan, Moreno, and
  Vasconcelos]{carneiro07}
G.~Carneiro, A.B. Chan, P.J. Moreno, and N.~Vasconcelos.
\newblock Supervised learning of semantic classes for image annotation and
  retrieval.
\newblock \emph{IEEE Transactions on Pattern Analysis and Machine
  Intelligence}, 29\penalty0 (3):\penalty0 394--410, 2007.

\bibitem[Chan et~al.(2010{\natexlab{a}})Chan, Coviello, and
  Lanckriet]{Chan2010tr}
A.~B. Chan, E.~Coviello, and G.R.G. Lanckriet.
\newblock Derivation of the hierarchical {EM} algorithm for dynamic textures.
\newblock Technical report, City University of Hong Kong, 2010{\natexlab{a}}.

\bibitem[Chan et~al.(2010{\natexlab{b}})Chan, Coviello, and
  Lanckriet]{Chan2010cvpr}
A.B. Chan, E.~Coviello, and G.R.G. Lanckriet.
\newblock {Clustering Dynamic Textures with the Hierarchical EM Algorithm}.
\newblock In \emph{Intl. Conference on Computer Vision and Pattern
  Recognition}, 2010{\natexlab{b}}.

\bibitem[Coviello et~al.(2011)Coviello, Chan, and Lanckriet]{coviello2011}
E.~Coviello, A.B. Chan, and G.R.G. Lanckriet.
\newblock Time series models for semantic music annotation.
\newblock \emph{Audio, Speech, and Language Processing, IEEE Transactions on},
  5\penalty0 (19):\penalty0 1343--1359, 2011.

\bibitem[Coviello et~al.(2012{\natexlab{a}})Coviello, Chan, and
  Gert]{coviello2012variational}
E.~Coviello, A.B. Chan, and G.R.G. Gert.
\newblock {The variational hierarchical EM algorithm for clustering hidden
  Markov models}.
\newblock In \emph{Advances in neural information processing systems},
  2012{\natexlab{a}}.

\bibitem[Coviello et~al.(2012{\natexlab{b}})Coviello, Mumtaz, Chan, and
  Lanckriet]{coviello2012growing}
E.~Coviello, A.~Mumtaz, A.B. Chan, and G.R.G. Lanckriet.
\newblock {Growing a Bag of Systems Tree for fast and accurate classification}.
\newblock In \emph{Computer Vision and Pattern Recognition (CVPR), 2012 IEEE
  Conference on}, pages 1979--1986. IEEE, 2012{\natexlab{b}}.

\bibitem[Csisz et~al.(1984)Csisz, Tusn{\'a}dy, et~al.]{csisz1984information}
I.~Csisz, G.~Tusn{\'a}dy, et~al.
\newblock Information geometry and alternating minimization procedures.
\newblock \emph{Statistics and decisions}, 1984.

\bibitem[Dempster et~al.(1977)Dempster, Laird, and Rubin]{Dempster1977}
A.~P. Dempster, N.~M. Laird, and D.~B. Rubin.
\newblock Maximum likelihood from incomplete data via the {EM} algorithm.
\newblock \emph{Journal of the Royal Statistical Society B}, 39:\penalty0
  1--38, 1977.

\bibitem[Doretto et~al.(2003)Doretto, Chiuso, Wu, and Soatto]{Doretto2003}
G.~Doretto, A.~Chiuso, Y.~N. Wu, and S.~Soatto.
\newblock Dynamic textures.
\newblock \emph{Intl. J. Computer Vision}, 51\penalty0 (2):\penalty0 91--109,
  2003.

\bibitem[Eck et~al.(2008)Eck, Lamere, Bertin-Mahieux, and Green]{eck08}
D.~Eck, P.~Lamere, T.~Bertin-Mahieux, and S.~Green.
\newblock Automatic generation of social tags for music recommendation.
\newblock In \emph{Advances in Neural Information Processing Systems}, 2008.

\bibitem[Frank and Asuncion(2010)]{Frank+Asuncion:2010}
A.~Frank and A.~Asuncion.
\newblock {UCI} machine learning repository, 2010.
\newblock URL \url{http://archive.ics.uci.edu/ml}.

\bibitem[Hall et~al.(1995)Hall, Horowitz, and Jing]{hall1995blocking}
P.~Hall, J.L. Horowitz, and B.Y. Jing.
\newblock On blocking rules for the bootstrap with dependent data.
\newblock \emph{Biometrika}, 82\penalty0 (3):\penalty0 561--574, 1995.

\bibitem[Hershey et~al.(2008)Hershey, Olsen, and
  Rennie]{hershey2008variational}
J.R. Hershey, P.A. Olsen, and S.J. Rennie.
\newblock {Variational Kullback-Leibler divergence for hidden Markov models}.
\newblock In \emph{Automatic Speech Recognition \& Understanding, 2007. ASRU.
  IEEE Workshop on}, pages 323--328. IEEE, 2008.

\bibitem[Hoffman et~al.(2009)Hoffman, Blei, and Cook]{hoffman09}
M.~Hoffman, D.~Blei, and P.~Cook.
\newblock Easy as {CBA}: A simple probabilistic model for tagging music.
\newblock In \emph{Proc. ISMIR}, pages 369--374, 2009.

\bibitem[Huang and Jack(1989)]{huang1989semi}
XD~Huang and MA~Jack.
\newblock {Semi-continuous hidden Markov models for speech signals}.
\newblock \emph{Computer Speech \& Language}, 3\penalty0 (3):\penalty0
  239--251, 1989.

\bibitem[Hubert and Arabie(1985)]{hubert1985comparing}
L.~Hubert and P.~Arabie.
\newblock Comparing partitions.
\newblock \emph{Journal of classification}, 2\penalty0 (1):\penalty0 193--218,
  1985.

\bibitem[Jaakkola(2000)]{Jaakkola00tutorialon}
Tommi~S. Jaakkola.
\newblock {Tutorial on Variational Approximation Methods}.
\newblock In \emph{In Advanced Mean Field Methods: Theory and Practice}, pages
  129--159. MIT Press, 2000.

\bibitem[Jebara et~al.(2004)Jebara, Kondor, and Howard]{jebara2004probability}
T.~Jebara, R.~Kondor, and A.~Howard.
\newblock Probability product kernels.
\newblock \emph{The Journal of Machine Learning Research}, 5:\penalty0
  819--844, 2004.

\bibitem[Jebara et~al.(2007)Jebara, Song, and Thadani]{jebara2007}
T.~Jebara, Y.~Song, and K.~Thadani.
\newblock {Spectral clustering and embedding with hidden Markov models}.
\newblock \emph{Machine Learning: ECML 2007}, pages 164--175, 2007.

\bibitem[Jordan et~al.(1999)Jordan, Ghahramani, Jaakkola, and
  Saul]{jordan1999introduction}
M.I. Jordan, Z.~Ghahramani, T.S. Jaakkola, and L.K. Saul.
\newblock An introduction to variational methods for graphical models.
\newblock \emph{Machine learning}, 37\penalty0 (2):\penalty0 183--233, 1999.

\bibitem[Juang and Rabiner(1985)]{juang1985}
B.~H. Juang and L.~R. Rabiner.
\newblock {A probabilistic distance measure for hidden Markov models}.
\newblock \emph{AT\&T Technical Journal}, 64\penalty0 (2):\penalty0 391--408,
  February 1985.

\bibitem[Kaufman and Rousseeuw(1987)]{rousseeuw1987clustering}
L.~Kaufman and P.~Rousseeuw.
\newblock Clustering by means of medoids.
\newblock \emph{Statistical data analysis based on the L1-norm and related
  methods}, pages 405--416, 1987.

\bibitem[Kleiner et~al.(2011)Kleiner, Talwalkar, Sarkar, and
  Jordan]{kleinerbootstrapping}
A.~Kleiner, A.~Talwalkar, P.~Sarkar, and M.I. Jordan.
\newblock Bootstrapping big data.
\newblock In \emph{Advances in Neural Information Processing Systems, Workshop:
  Big Learning: Algorithms, Systems, and Tools for Learning at Scale}, 2011.

\bibitem[Krogh et~al.(1994)Krogh, Brown, Mian, Sjolander, and
  Haussler]{krogh1994hidden}
A.~Krogh, M.~Brown, I.S. Mian, K.~Sjolander, and D.~Haussler.
\newblock {Hidden Markov models in computational biology. Applications to
  protein modeling}.
\newblock \emph{Journal of Molecular Biology}, 235\penalty0 (5):\penalty0
  1501--1531, 1994.

\bibitem[Lyngso et~al.(1999)Lyngso, Pedersen, and Nielsen]{lyngso1999metrics}
RB~Lyngso, CN~Pedersen, and H.~Nielsen.
\newblock {Metrics and similarity measures for hidden Markov models}.
\newblock In \emph{Proc. Int. Conf. Intell. Syst. Mol. Biol}, pages 178--186,
  1999.

\bibitem[Mandel and Ellis(2008)]{mandel08}
M.I. Mandel and D.P.W. Ellis.
\newblock {Multiple-instance learning for music information retrieval}.
\newblock In \emph{Proc. ISMIR}, pages 577--582, 2008.

\bibitem[Nag et~al.(1986)Nag, Wong, and Fallside]{nag1986script}
R.~Nag, K.~Wong, and F.~Fallside.
\newblock {Script recognition using hidden Markov models}.
\newblock In \emph{Acoustics, Speech, and Signal Processing, IEEE International
  Conference on ICASSP'86.}, volume~11, pages 2071--2074. IEEE, 1986.

\bibitem[Neal and Hinton(1998)]{neal1998view}
R.M. Neal and G.E. Hinton.
\newblock {A view of the EM algorithm that justifies incremental, sparse, and
  other variants}.
\newblock \emph{NATO ASI Series D Behavioral and Social Sciences}, 89:\penalty0
  355--370, 1998.

\bibitem[Panuccio et~al.(2002)Panuccio, Bicego, and Murino]{panuccio2002hidden}
A.~Panuccio, M.~Bicego, and V.~Murino.
\newblock {A hidden Markov model-based approach to sequential data clustering}.
\newblock \emph{Structural, Syntactic, and Statistical Pattern Recognition},
  pages 734--743, 2002.

\bibitem[Penny and Roberts(2000)]{penny2000notes}
WD~Penny and SJ~Roberts.
\newblock Notes on variational learning.
\newblock Technical report, Technical report, Oxford University, 2000.

\bibitem[Qi et~al.(2007)Qi, Paisley, and Carin]{qi2007music}
Y.~Qi, J.W. Paisley, and L.~Carin.
\newblock {Music analysis using hidden Markov mixture models}.
\newblock \emph{Signal Processing, IEEE Transactions on}, 55\penalty0
  (11):\penalty0 5209--5224, 2007.

\bibitem[Rabiner and Juang(1993)]{Rabiner93}
L.~Rabiner and B.~H. Juang.
\newblock \emph{{Fundamentals of Speech Recognition}}.
\newblock Prentice Hall, {Upper Saddle River (NJ, USA)}, 1993.

\bibitem[Reed and Lee(2006)]{reed06}
J.~Reed and C.H. Lee.
\newblock {A study on music genre classification based on universal acoustic
  models}.
\newblock In \emph{Proc. ISMIR}, pages 89--94, 2006.

\bibitem[Scaringella and Zoia(2005)]{scaringella2005modeling}
N.~Scaringella and G.~Zoia.
\newblock On the modeling of time information for automatic genre recognition
  systems in audio signals.
\newblock In \emph{Proc. ISMIR}, pages 666--671, 2005.

\bibitem[Smyth(1997)]{smyth1997}
P.~Smyth.
\newblock {Clustering sequences with hidden Markov models}.
\newblock In \emph{Advances in neural information processing systems}, 1997.

\bibitem[Theodoridis and Hu(2007)]{theodoridis2007action}
T.~Theodoridis and H.~Hu.
\newblock {Action classification of 3d human models using dynamic ANNs for
  mobile robot surveillance}.
\newblock In \emph{Robotics and Biomimetics, 2007. ROBIO 2007. IEEE
  International Conference on}, pages 371--376. IEEE, 2007.

\bibitem[Turnbull et~al.(2008)Turnbull, Barrington, Torres, and
  Lanckriet]{music:turnbull08}
D.~Turnbull, L.~Barrington, D.~Torres, and G.~Lanckriet.
\newblock Semantic annotation and retrieval of music and sound effects.
\newblock \emph{IEEE Transactions on Audio, Speech and Language Processing},
  16\penalty0 (2):\penalty0 467--476, February 2008.

\bibitem[Vasconcelos(2001)]{Vasc2001b}
N.~Vasconcelos.
\newblock Image indexing with mixture hierarchies.
\newblock In \emph{IEEE Conf. Computer Vision and Pattern Recognition}, 2001.

\bibitem[Vasconcelos and Lippman(1998)]{Vasc1998}
N.~Vasconcelos and A.~Lippman.
\newblock Learning mixture hierarchies.
\newblock In \emph{Advances in Neural Information Processing Systems}, 1998.

\bibitem[Wainwright and Jordan(2008)]{wainwright2008graphical}
M.J. Wainwright and M.I. Jordan.
\newblock Graphical models, exponential families, and variational inference.
\newblock \emph{Foundations and Trends in Machine Learning}, 1\penalty0
  (1-2):\penalty0 1--305, 2008.

\bibitem[Williams et~al.(2006)Williams, Toussaint, and
  Storkey]{williams2006extracting}
B.~Williams, M.~Toussaint, and A.~Storkey.
\newblock Extracting motion primitives from natural handwriting data.
\newblock \emph{Artificial Neural Networks--ICANN 2006}, pages 634--643, 2006.

\bibitem[Yin and Yang(2005)]{yin2005integrating}
J.~Yin and Q.~Yang.
\newblock {Integrating hidden Markov models and spectral analysis for sensory
  time series clustering}.
\newblock In \emph{Data Mining, Fifth IEEE International Conference on}, pages
  8--pp. IEEE, 2005.

\bibitem[Zhong and Ghosh(2003)]{zhong2003unified}
S.~Zhong and J.~Ghosh.
\newblock {A unified framework for model-based clustering}.
\newblock \emph{The Journal of Machine Learning Research}, 4:\penalty0
  1001--1037, 2003.

\end{thebibliography}

\end{document}